\documentclass{article}
\usepackage[usenames,dvipsnames,dvinames]{xcolor}

\usepackage[accepted]{icml2019}

\usepackage{graphicx}
\usepackage{amssymb,amsmath,bm}
\usepackage[utf8]{inputenc} 
\usepackage[T1]{fontenc}    
\usepackage{hyperref}       
\usepackage{url}            
\usepackage{booktabs}       
\usepackage{amsfonts}       
\usepackage{nicefrac}       
\usepackage{microtype}      
\usepackage{color}
\usepackage{subcaption}
\usepackage{wrapfig}
\usepackage{float}
\usepackage{multirow}
\usepackage{tabularx}
\usepackage{ifthen}

\usepackage{titlesec}
\titlespacing\section{0pt}{4pt plus 2pt minus 2pt}{4pt plus 2pt minus 2pt}
\titlespacing\subsection{0pt}{3pt plus 2pt minus 2pt}{3pt plus 2pt minus 2pt}
\titlespacing\subsubsection{0pt}{2pt plus 1pt minus 1pt}{2pt plus 1pt minus 1pt}

\usepackage{amsthm}
\newtheorem{thm}{Theorem}
\newtheorem{lem}[thm]{Lemma}

\newenvironment{sproof}{%
	\proof}{\endproof}

\icmltitlerunning{Combating Label Noise in Deep Learning Using Abstention}

\begin{document}
	
	\twocolumn[
		\icmltitle{Combating Label Noise in Deep Learning Using Abstention}
		\title{Combating Label Noise in Deep Learning Using Abstention}
		

%





\begin{icmlauthorlist}
	\icmlauthor{Sunil Thulasidasan}{lanl,uw}
	\icmlauthor{Tanmoy Bhattacharya}{lanl}
	\icmlauthor{Jeffrey Bilmes}{uw} 
	\icmlauthor{Gopinath Chennupati}{lanl}
	\icmlauthor{Jamaludin Mohd-Yusof}{lanl}
\end{icmlauthorlist}


\icmlaffiliation{lanl}{Los Alamos National Laboratory, Los Alamos, NM, USA}
\icmlaffiliation{uw}{Department of Electrical \& Computer Engineering, University of Washington, Seattle, WA, USA}

\icmlcorrespondingauthor{Sunil Thulasidasan}{sunil@lanl.gov}


\vskip 0.3in
]
\newcommand{\fix}{\marginpar{FIX}}
\newcommand{\new}{\marginpar{NEW}}

\providecommand{\dodraft}{true}
\newboolean{isdraft}
\setboolean{isdraft}{\dodraft} 
\ifthenelse{\boolean{isdraft}}{%
	\newcommand{\sunil}[2]{{\color{blue}{#1 $\to$ {\bf Sunil}}: $\lll$ #2 $\rrr$}}
	\newcommand{\jeff}[2]{{\color{red}{#1 $\to$ {\bf Jeff}}: $\lll$ #2 $\rrr$}}
	\newcommand{\todo}[1]{{\color{red}{{\bf TODO: #1} }}}
	\newcommand{\fixme}[1]{{\color{red}{{\bf FIXME: $\lll$ #1 $\ggg$} }}}
        \newcommand{\clarifyme}[1]{{\color{OliveGreen}{{\bf CLARIFY: $\lll$ #1 $\ggg$} }}}
	\newcommand{\afterfix}[1]{{\color{red}{{\bf AFTERFIX: #1} }}}
	
} {
	
	\newcommand{\sunil}[2]{}
	\newcommand{\jeff}[2]{}
	\newcommand{\todo}[1]{}
	\newcommand{\fixme}[1]{}
        \newcommand{\clarifyme}[1]{}
	\newcommand{\afterfix}[1]{}
	
}

\printAffiliationsAndNotice{}
\begin{abstract}

We introduce a novel method to combat label noise when training deep neural networks for classification. 
 We propose a  loss function that permits  abstention during training thereby allowing the  DNN to abstain on confusing samples  while continuing to learn and improve classification performance on the non-abstained samples. We show how such a deep abstaining classifier (DAC) can be used for robust learning in the presence of different types of label noise. In the case of structured or systematic label noise -- where noisy training labels or confusing examples are correlated with underlying features  of the data-- training with abstention enables representation learning for features that are associated with unreliable labels. In the case of unstructured (arbitrary) label noise, abstention during training enables the  DAC to be used as an effective data cleaner by identifying samples that are likely to have label noise.
   We provide analytical results on the loss function behavior that enable dynamic adaption of  abstention rates based on learning progress during training. We demonstrate the utility of the deep abstaining classifier for various image classification tasks under different types of label noise; in the case of arbitrary label noise, we show significant improvements over previously published results on multiple image benchmarks. Code is available at \href{https://github.com/thulas/dac-label-noise}{https://github.com/thulas/dac-label-noise}

\end{abstract}

\section{Introduction}



The impressive performance of deep neural networks in recent years in various tasks such as image classification and speech recognition~\cite{lecun2015deep} have been made possible by the availability of large quantities of human-annotated (i.e., labeled) training data. For example, the ImageNet database~\citep{deng2009imagenet} used for training vision classifiers has now over 14 million hand annotated  images\citep{imagenet_stats}. Though enabling deep models to match or surpass human performance, the requirement of vast quantities of human-annotated datasets can often be a bottleneck in training deep learning systems. There are generally two approaches to tackle this problem: use a scalable and affordable annotation platform like Amazon Mechanical Turk~\citep{amturk} or alternatively, automatically collect large amounts of web-based data that have associated meta-information (tags, for instance) which are used for labeling. In both cases, a certain amount of erroneously labeled data is bound to occur, often in significant fractions for the latter situation~\citep{li2017webvision}. While there is empirical evidence that deep networks are robust to some amount of label noise~\cite{rolnick2017deep}, significant label noise can degrade generalization performance  due to the ability of deep models to fit random labels~\citep{zhang2016understanding}. 
In such situations, it is often better to eliminate the noisy data and train with just the cleaner subset~\cite{frenay2014classification}, or use a semi-supervised approach~\citep{chapelle2009semi} which uses all the samples, but only retains the labels in the cleaner set for training. In both cases, one needs to identify the subset of training data whose labels are likely to be unreliable.

While label noise is a well studied problem in machine learning~\citep{nettleton2010study,zhu2004class,sukhbaatar2014training,reed2014training,patrini2017making}, there has not been much work in identifying and ignoring it during the process of training itself.
In this paper we propose a novel {\em abstention} (or rejection) based mechanism to combat label noise while training  deep models. 
Most of the  theoretical and empirical investigations into rejection classification systems have been studied in a post-processing setting -- i.e., a classifier is first trained as usual, and an abstention threshold is determined based on post-training performance on a calibration set; the DNN then abstains on uncertain predictions during inference. Abstention classifiers have been proposed for shallow learners~\citep{chow1970optimum,cortes2016learning,fumera2002support} and for multilayer perceptrons~\citep{de2000reject}.
In the context of deep networks, this has been an under-explored area with~\citep{geifman2017selective} recently proposing an effective technique of selective classification for optimizing risk-vs-coverage profiles based on the output of a trained model.

In contrast to the above, the approach described in this paper 
  employs abstention during {\em training} as well as inference. This gives the DNN an option to abstain on a confusing 
training sample thereby mitigating the misclassification loss but incurring an abstention penalty. We empirically show that our formulation ensures that the DNN continues to learn the true class even while abstaining, progressively reducing its abstention rate as it learns on the true classes and finally abstaining on only the most confusing samples. We demonstrate the advantages of such a formulation under two different label-noise scenarios: 
first, when labeling errors are correlated with some underlying feature of the data (systematic or structured label noise), abstention training allows the DNN to learn features that are indicative of unreliable training signals  which are thus likely to lead to uncertain predictions. This kind of representation learning for abstention is useful both for effectively eliminating structured noise and also for interpreting  the reasons for abstention. 
 Second, 
%
 we  show how an abstention-based approach can be used as an effective {\em data cleaner} when training data contains arbitrary (or unstructured) label noise. A DNN trained with an abstention option can be used to identify and filter out noisy training data leading to significant performance benefits for downstream training using a cleaner set.
To summarize, the contributions of this paper are:

\begin{itemize}
	\item The introduction of abstention-based training as an effective approach to combat label noise.We show that such a deep abstaining classifier (DAC)  enables robust learning in the presence of label noise.
		
	\item The demonstration of the ability of the DAC to learn features associated with systematic label noise. Through numerous experiments, we show how the DAC is able to pick up (and then abstain on) such features with high precision.
	
	\item Demonstration of the utility of the  DAC as an effective {\em data cleaner } in the presence of arbitrary label noise. We provide results on learning with noisy labels on multiple image benchmarks (CIFAR-10, CIFAR-100 and Fashion-MNIST) that improve upon existing methods. Our method is also considerably simpler to implement and can be used with any existing DNN architecture as only the loss function is changed.
		
\end{itemize}

While ideally such an abstaining classifier should also learn to reliably abstain when presented with adversarially perturbed samples~\citep{nguyen2015deep,szegedy2013intriguing, moosavi2017universal}, in this work we do not consider adversarial settings and leave that for future exploration. The rest of the paper is organized as follows: Section~\ref{sec:loss_formulation} describes the loss function formulation and an algorithm for automatically tuning abstention behavior. Section~\ref{sec:struct_noise} discusses learning in the presence of structured noise, including experimental results and visual interpretations of abstention. Section~\ref{sec:unstructured_noise} presents the utility of the DAC for data cleaning in the presence of unstructured (arbitrary) noise. Section~\ref{sec:abst_memorization} has further discussions on abstention behavior in the context of memorization. 
  We conclude in Section~\ref{sec:conclusion}. 

\section{Loss Function for the Deep Abstaining Classifier}
\label{sec:loss_formulation}


We assume we are interested in training a $k$-class multi-class classifier with a
deep neural network (DNN) where $x$ is the input and $y$ is the
output. For a given $x$, we define $p_i = p_w(y=i|x)$ (the probability
of the $i\text{th}$ class given $x$) as the $i^\text{th}$ output of the DNN that
implements the probability model $p_w(y=i|x)$ where $w$ is the set of
weight matrices of the DNN. For notational brevity, we use $p_i$ in
place of $p_w(y=i|x)$ when the input context $x$ is clear.

The standard cross-entropy training loss for DNNs then takes the form
$\mathcal L_\text{standard} = -\sum_{i=1}^k t_i \log p_i$ where $t_i$
is the target for the current sample.  The DAC has an additional $k+1^\text{st}$ output $p_{k+1}$ which is
meant to indicate the probability of abstention. We train
the DAC with following modified version of the $k$-class
cross-entropy per-sample loss:
\small
\begin{equation}
  \mathcal{L}(x_j) =
  ( 1 - p_{k+1})\left(-\sum_{i=1}^k t_i \log \frac{p_i}{1-p_{k+1}}\right)
  + \alpha \log \frac{1} { 1 - p_{k+1} } \label{eqn:loss_fn}
\end{equation}
\normalsize
The first term is a modified cross-entropy loss over the $k$
non-abstaining classes. Absence of the abstaining output (i.e.,
$p_{k+1}=0$) recovers exactly the usual cross-entropy; otherwise, the
abstention mass has been normalized out of the $k$ class probabilities.
The second term penalizes abstention and is weighted by
$\alpha \geq 0$, a hyperparameter expressing the degree of penalty.
If $\alpha$ is very large, there is a high penalty for abstention 
driving $p_{k+1}$ to zero and recovering the standard unmodified
cross-entropy loss; in such case, the model learns to never abstain.
With $\alpha$ very small, the classifier may abstain on everything
with impunity since the adjusted cross-entropy loss becomes zero and
it does not matter what the classifier does on the $k$ class
probabilities.  When $\alpha$ is between these extremes, things become
more interesting: whether the DNN chooses to abstain or not depends on how much cross-entropy error it is making while learning on the true classes; it is this error that drives mass into the abstention class subject to the abstention penalty hyperparameter $\alpha$.


\begin{lem}
	For the loss function $\mathcal{L}$ given in Equation~\ref{eqn:loss_fn}, if $j$ is the true class for sample $x$, then as long as $\alpha \geq 0$, $\frac{\partial \mathcal{L} }{\partial a_{j}} \leq 0$ (where $a_j$ is the pre-activation into the softmax unit of class $j$).
\end{lem}

The proof is given in Section A 
of the supplementary material.  This ensures that, during gradient descent, learning on the true classes persists even in the presence of abstention, even though the true class might not end up be the winning class. We provide additional discussion on abstention behavior in Section~\ref{sec:abst_memorization}.
\subsection{Auto-tuning $\alpha$}
\label{sec:alpha_auto}
Let $g = -\sum_{i=1}^k t_i \log p_i $ be the standard cross-entropy
loss and $a_{k+1}$ be the pre-activation into the softmax unit for the
abstaining class. Then it is easy to see that:
\small
\begin{equation}
\frac{\partial \mathcal{L} }{\partial a_{k+1}} = p_{k+1}\left[ \left(1 - p_{k+1}\right) \left [ \log \frac{1}{1-p_{k+1}} -g \right] + \alpha \right ].
\end{equation}
\normalsize
%
%
%
%
%
%
%

\begin{algorithm}[tb]
	\caption{$\alpha$ auto-tuning}
	\label{alg:auto_tune}
	\begin{algorithmic}	
		\STATE {\bfseries Input:} total iter. ($T$), current iter. ($t$), total epochs ($E$), abstention-free epochs   ($L$), current epoch ($e$),  $\alpha$ init factor ($\rho$), final $\alpha$ ($\alpha_{final}$), mini-batch cross-entropy over true classes ($\mathcal{H}_c(P^M_{1\dots K}$)\\
		\STATE $\alpha_{set} = False$
		\FOR {$t := 0$ to $T$}
			\IF{$e < L$} 
				\STATE $\beta = (1-{P_{k+1}^M}) \mathcal{H}_c(P^M_{1\dots K})$ 
					\IF{$t = 0$} 
						\STATE $\tilde{\beta} = \beta$\COMMENT{ // initialize moving average}
					\ENDIF
					\STATE $\tilde{\beta} \leftarrow (1-\mu) \tilde{\beta} + \mu \beta$
			\ENDIF
			\IF{ $e = L$ and  {\bf not} $\alpha_{set}$}
				\STATE $\alpha := \tilde{\beta}/\rho$\COMMENT{ // initialize $\alpha$ at start of epoch $L$}
				\STATE $\delta_{\alpha} := \frac{\alpha_{final} - \alpha }{E-L}$ 
				\STATE $update_{epoch} = L$
				\STATE $\alpha_{set} = True$
			\ENDIF
			\IF { $e  >  update_{epoch}$ }
				\STATE $\alpha \leftarrow \alpha + \delta_{\alpha}$\COMMENT{ //then update $\alpha$ once every epoch}
				\STATE $update_{epoch} = e$
			\ENDIF
		\ENDFOR
	\end{algorithmic}
\end{algorithm}

During gradient descent, abstention pre-activation is increased if $\frac{\partial \mathcal{L} }{\partial a_{k+1}} < 0$. 
The threshold on $\alpha$ for this is $\alpha < \left( 1 - p_{k+1} \right) \left (- \log \frac{p_j}{1-p_{k+1}}\right )$
where $j$ is the true class for sample $x$.  If only a small fraction of the mass over the actual classes is in the true class $j$, then the DAC has not learned to correctly classify that particular sample from class $j$, and will push mass into the abstention class provided $\alpha$ satisfies the above inequality. This constraint allows us to perform auto-tuning on $\alpha$ during training (Algorithm~\ref{alg:auto_tune}). 
$\tilde{\beta}$ is a smoothed moving average of the $\alpha$ threshold (initialized to 0), and updated at every mini-batch iteration. We perform abstention-free training for $L$ initial epochs (a warm-up period) to accelerate learning, triggering abstention from epoch $L+1$ onwards. At the start of abstention, $\alpha$ is initialized to a much smaller value than the threshold $\tilde{\beta}$  to encourage abstention on all but the easiest of examples learnt so far. As the learning progresses on the true classes, abstention is reduced. We linearly ramp up $\alpha$ over the remaining epochs (updating once per epoch )to a final value of $\alpha_{final}$ In the experiments in the subsequent sections, we illustrate how the DAC, when trained with this loss function, learns representations for abstention remarkably well.

\section{The DAC as a Learner of Structured Noise}
\label{sec:struct_noise}
While noisy training labels are usually an unavoidable  occurrence in real-world data, such  noise can often exhibit a pattern attributable to  training data being corrupted in some non-arbitrary or systematic manner. This kind of label noise  can occur when some classes are more likely to be mislabeled than others, either  because of confusing features or a lack of sufficient level of expertise or unreliability of the annotator. For example, the occurrence of non-iid, systematic  label noise in brain-computer interface applications -- where noisy data is correlated with the state of the participant --  has been documented in~\citep{porbadnigk2014brain, gornitz2014learning}. In image data collected for training (that might have been automatically pre-tagged by a recognition system), a subset of the  images might be of degraded quality, causing such labels to be unreliable\footnote{We assume, in this case, one only has access to the labels, and not the confidence scores}. Further, systematic noise can also occur if all the data were labeled using the same mechanism~\citep{brodley1999identifying}; for a comprehensive survey of label noise see~\citep{frenay2014classification}.



%

%
%


%

\begin{figure*}[htb]
	\centering
	\begin{subfigure}[b]{0.12\textwidth}
		\includegraphics[width=\columnwidth]{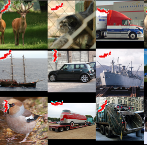}
		\caption{}
		\label{fig:smudged_sample}
	\end{subfigure}
	\begin{subfigure}[b]{0.2\textwidth}
		\includegraphics[width=\columnwidth]{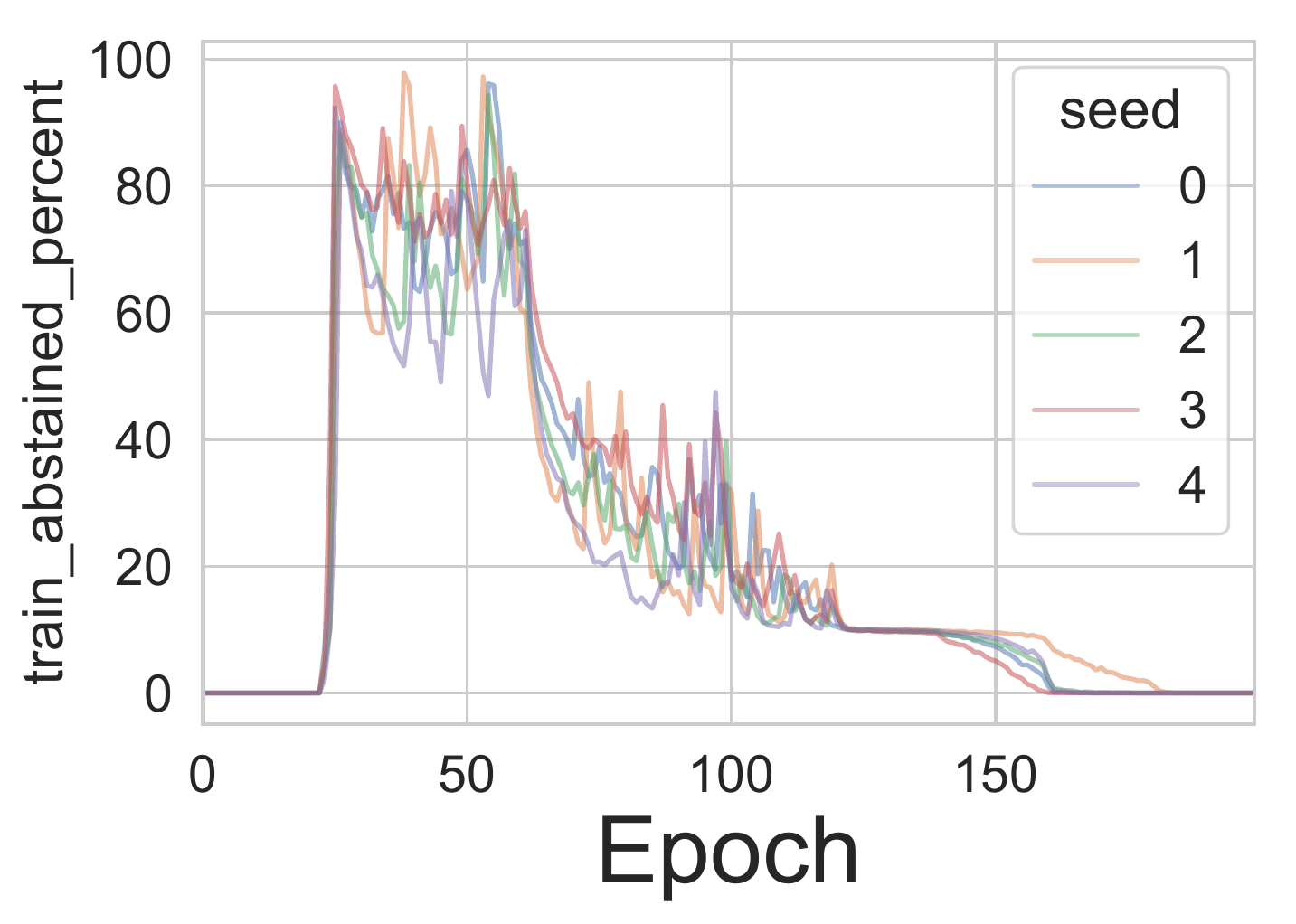}
		\caption{}
		\label{fig:smudged_train_abstain}
	\end{subfigure}
	\begin{subfigure}[b]{0.2\textwidth}
		\includegraphics[width=\columnwidth]{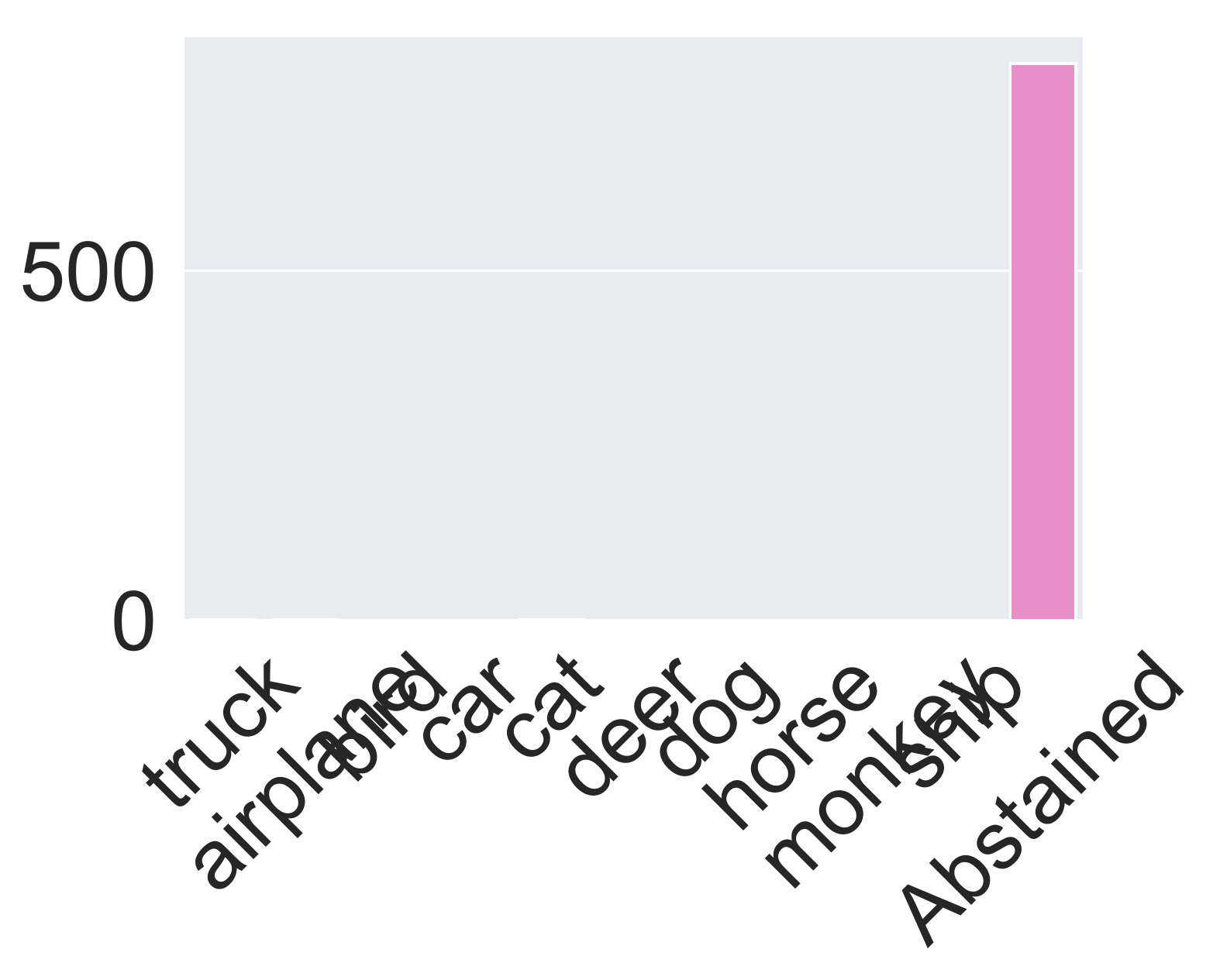}
		\caption{}
		\label{fig:dac_smudged_abst_recall}
	\end{subfigure}
	\begin{subfigure}[b]{0.2\textwidth}
		\includegraphics[width=\columnwidth]{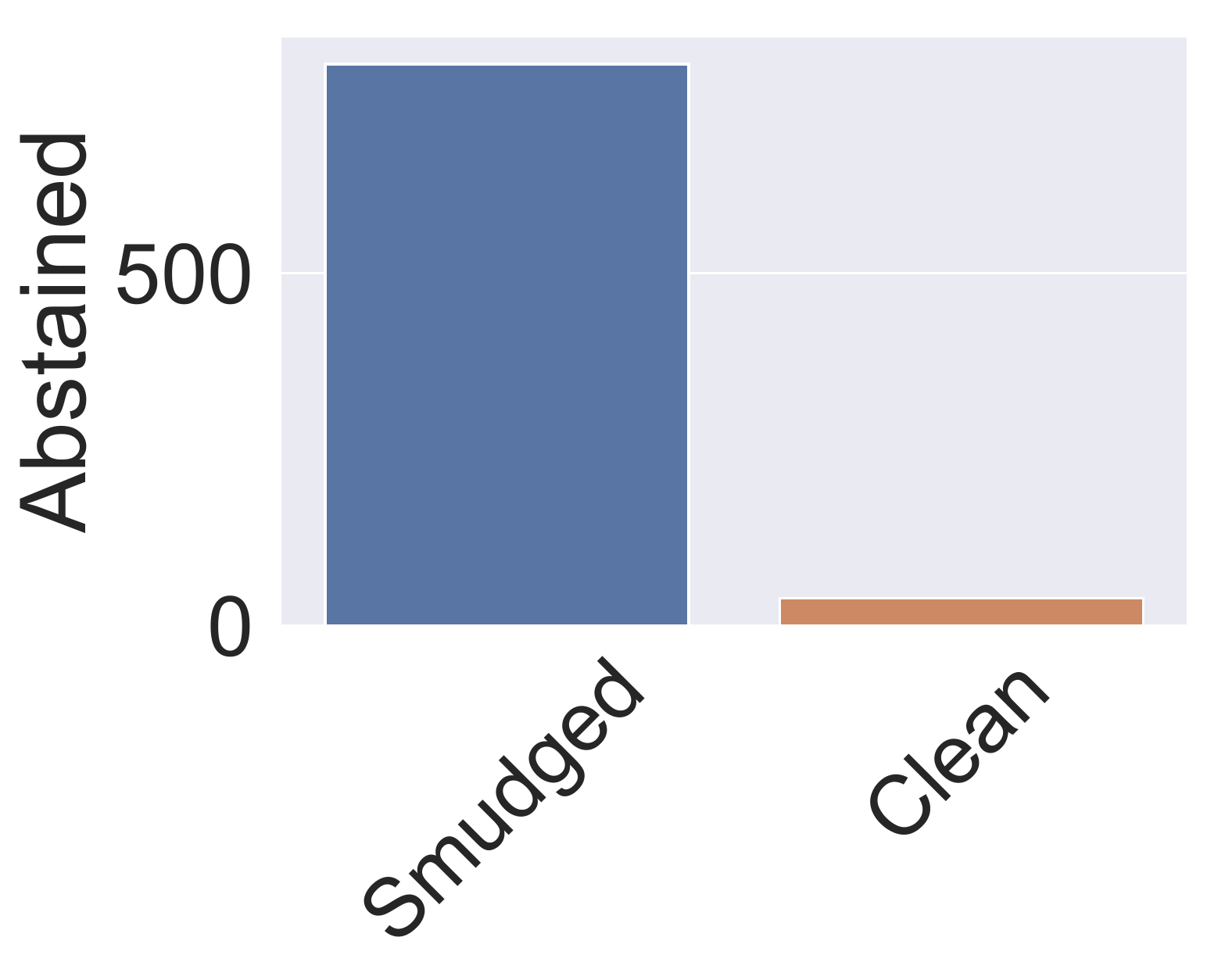}
		\caption{}
		\label{fig:dac_smudge_pred_dist}	
	\end{subfigure}
	\begin{subfigure}[b]{0.2\textwidth}
		\includegraphics[width=\columnwidth]{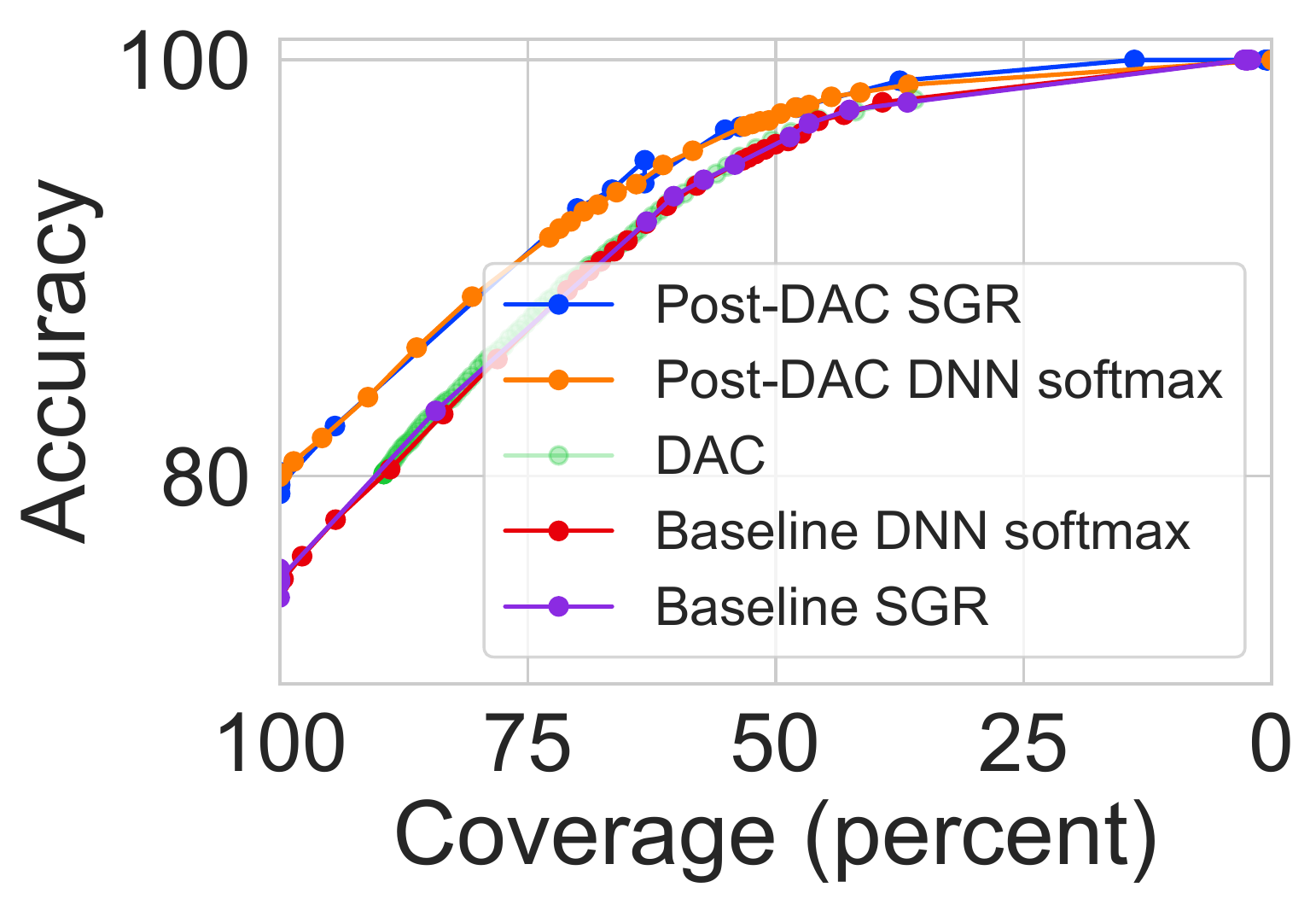}
		\caption{}
		\label{fig:smudge_dac_vs_dnn_softmax_threshold}
	\end{subfigure}

	\caption{(a) A sample of smudged images on which labels were randomized.(b) Abstention percentage on training set as training progresses (c)The DAC abstains on most of the smudged images on the test set (abstention recall) (d) Almost all of the abstained images were those that were smudged (abstention precision)  (e) Risk-coverage curves for baseline DNNs, DAC and post-DAC DNN. For the baseline and post-DAC DNNs, we report coverage based on softmax and SGR thresholds. First-pass training with the DAC improves performance for both softmax and SGR methods.}
	\label{fig:sample_of_smudged_images}
\end{figure*}

In  these scenarios, there are usually consistent indications in the input $x$ that
tend to be correlated with noise in the labels,
but such
correlations are rarely initially obvious. Given the large amount of
data required to train deep models, the process of curating the data down to a
clean, reliable set might be prohibitively expensive. In situations involving sensitive data (patient records, for example) crowd-sourcing label annotations might not even be an option.
However, given that DNNs can learn rich, hierarchical
representations
, one of the  questions we explore in this paper is whether we can exploit the representational power of DNNs
to {\em learn} such feature mappings that are indicative of unreliable
or confusing samples. Since abstention is driven by the cross-entropy in the training loss, features that are consistently picked up by the DAC during abstention should thus have high feature weights with respect to  the abstention class, suggesting that the DAC might learn to make such associations. In the following sections, 
we describe a series of experiments on image data that
demonstrate precisely this behavior -- using abstention training, the
DAC learns features that are associated with difficult or confusing
samples and reliably learns to abstain based on these features.


\subsection{Experiments}


{\bf Setup:} For the experiments in this section, 
 we use a deep convolutional network employing the VGG-16~\citep{simonyan2014very} architecture, implemented in the PyTorch~\citep{paszke2017automatic} framework. We train the network for 200 epochs using SGD accelerated with Nesterov momentum and employ a weight decay of $.0005$, initial learning rate of $0.1$ and learning rate annealing using an annealing factor of $0.5$ at epoch $60,120$ and $160$. We perform abstention-free training during the first 20 epochs which allows for faster training\footnote{Training with abstention from the start just means we have to train for a longer number of epochs to reach a given abstention-vs-accuracy point.} To enable better visualization, in this section, we use the labeled version of the STL-10 dataset~\citep{coates2011analysis}, comprising of 5000 and 8000 96x96 RGB images in the train and test set respectively, augmented with random crops and horizontal flips during training. We use this architecture and dataset combination to keep training times reasonable, but over a relatively challenging dataset with complex features. For the $\alpha$ auto-update algorithm we set $\rho$ ($\alpha$ initialization factor) to 64 and $\mu$ to 0.05; we did not tune these parameters. 


%
%

\subsection{Noisy Labels Co-Occurring with an Underlying Cross-Class Feature}
\label{sec:rand_smudge}
\begin{figure*}[htbp]
	\centering
	\begin{subfigure}[b]{0.2\textwidth}
		\includegraphics[width=\columnwidth]{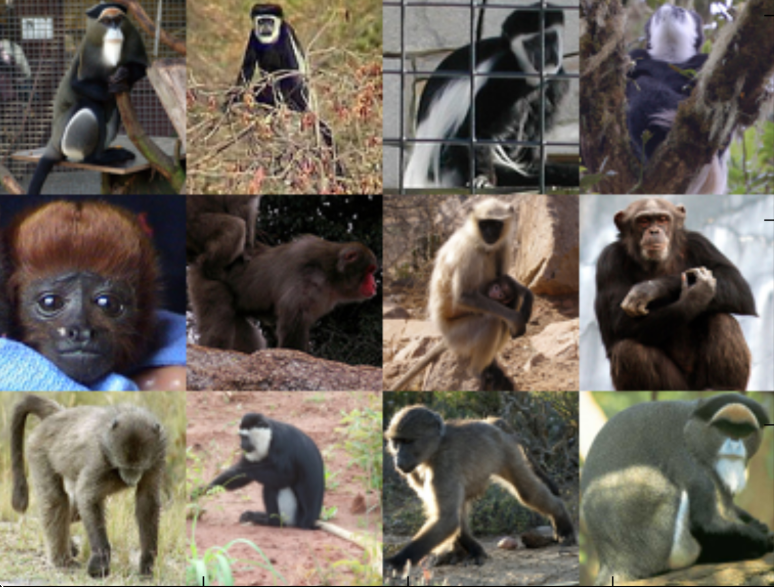}
		\caption{}
		\label{fig:monkey_sample}
	\end{subfigure}
	\begin{subfigure}[b]{0.2\textwidth}
		\includegraphics[width=\columnwidth]{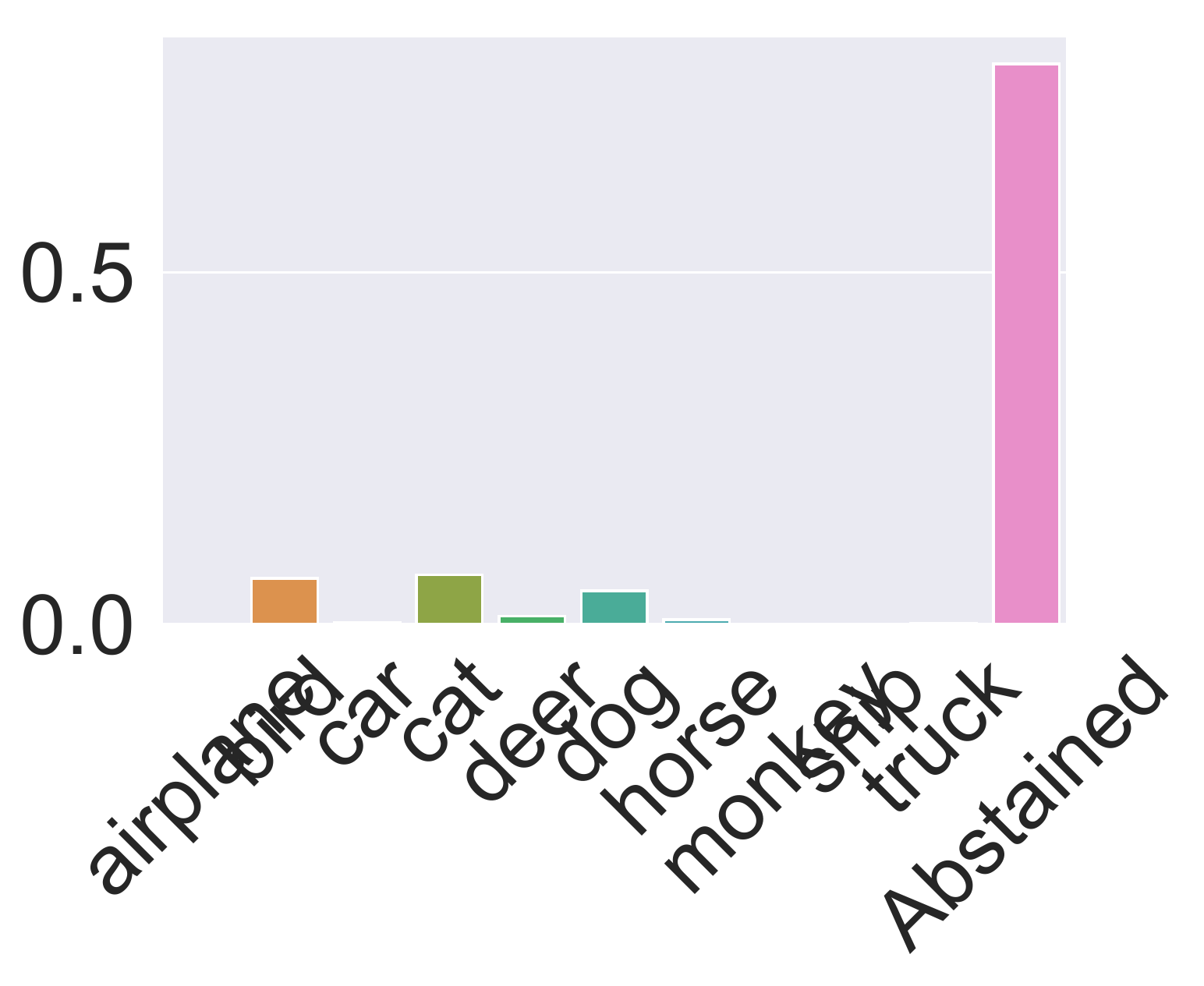}
		\caption{}
		\label{fig:dac_monkey_abst_recall}
	\end{subfigure}
	\begin{subfigure}[b]{0.2\textwidth}
		\includegraphics[width=\columnwidth]{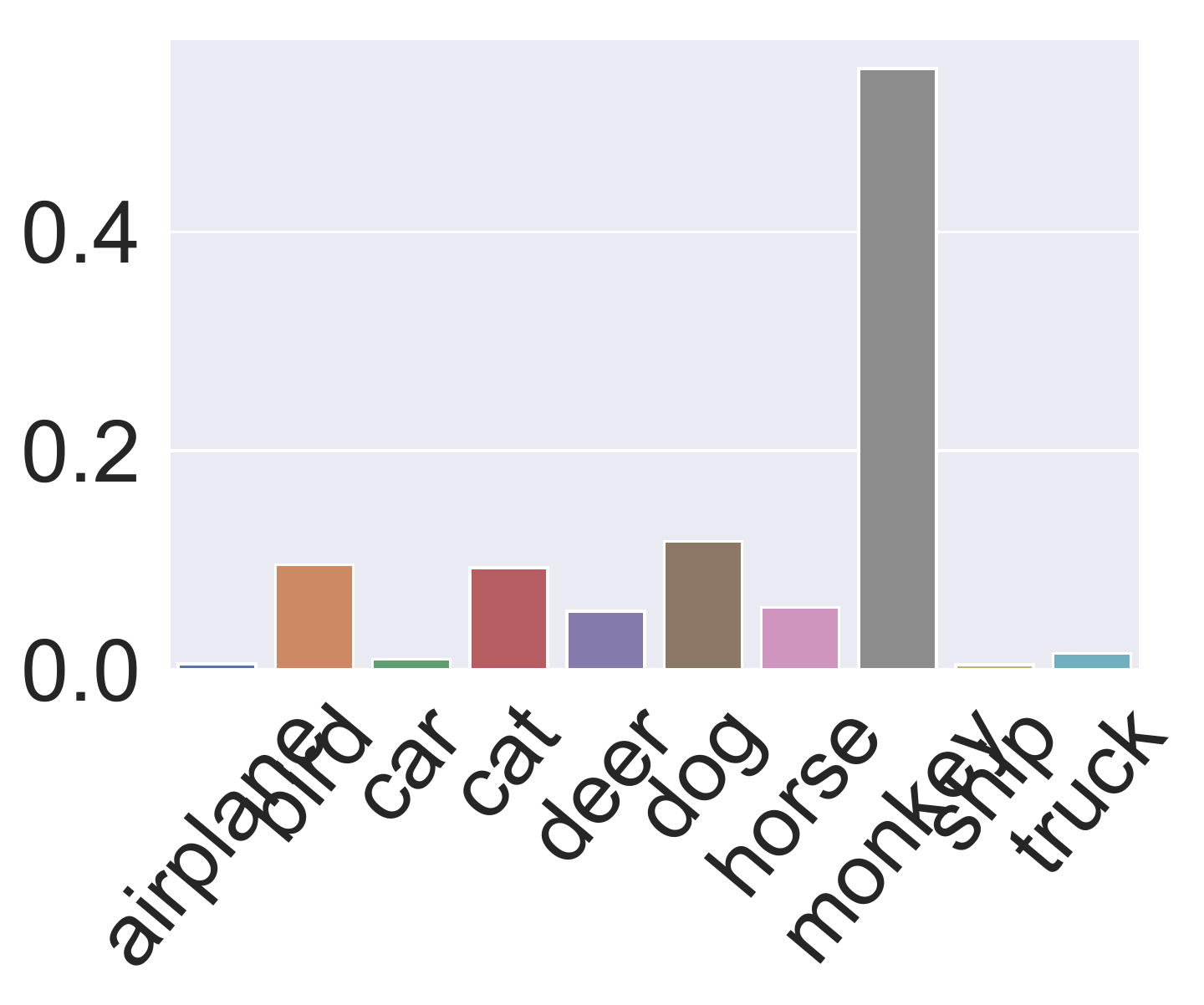}
		\caption{}
		\label{fig:dac_monkey_abst_precision}
	\end{subfigure}
	\begin{subfigure}[b]{0.2\textwidth}
		\includegraphics[width=\columnwidth]{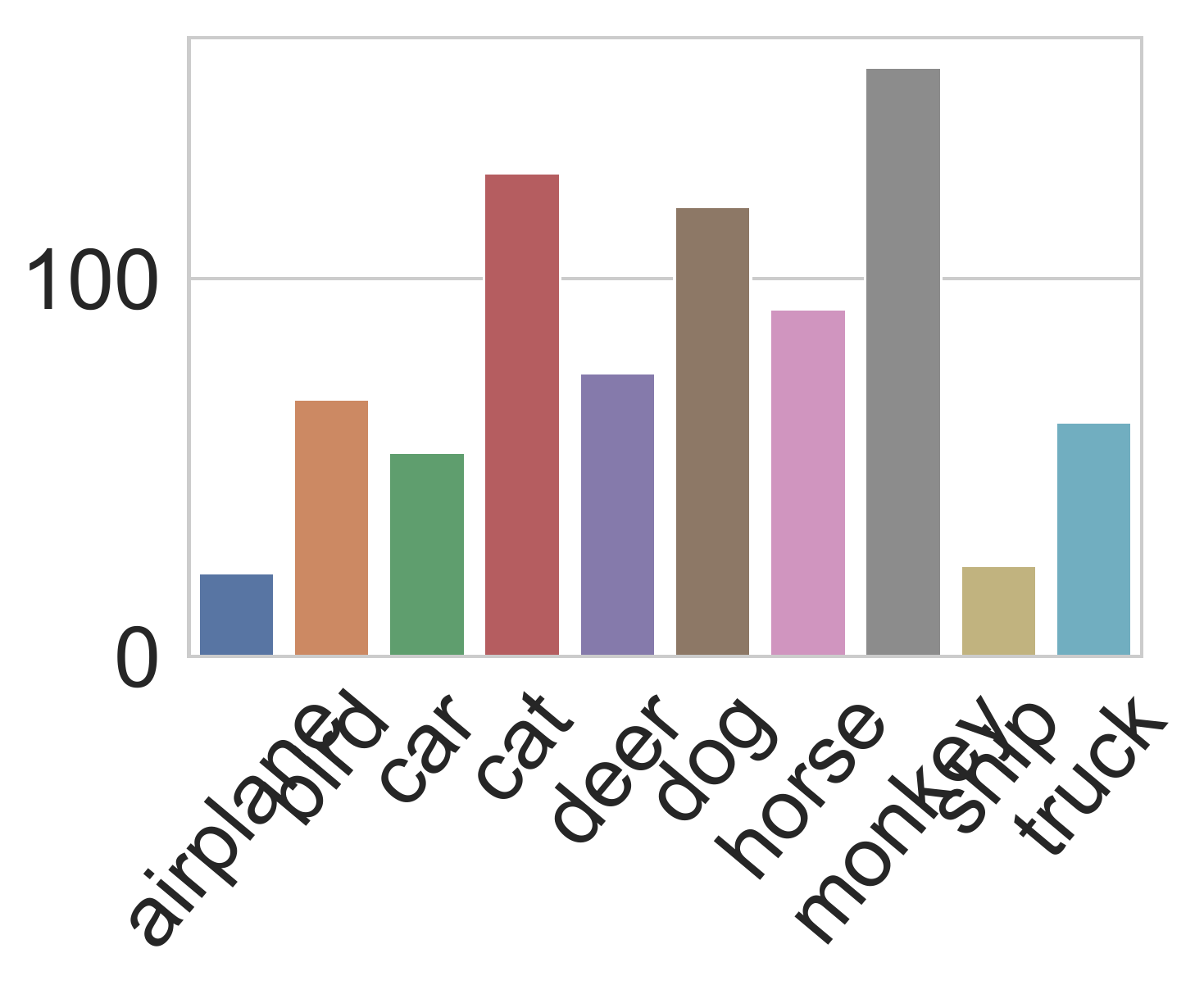}
		\caption{}
		\label{fig:dnn_monkey_dist}	
	\end{subfigure}
	\begin{subfigure}[b]{0.2\textwidth}
		\includegraphics[width=\columnwidth]{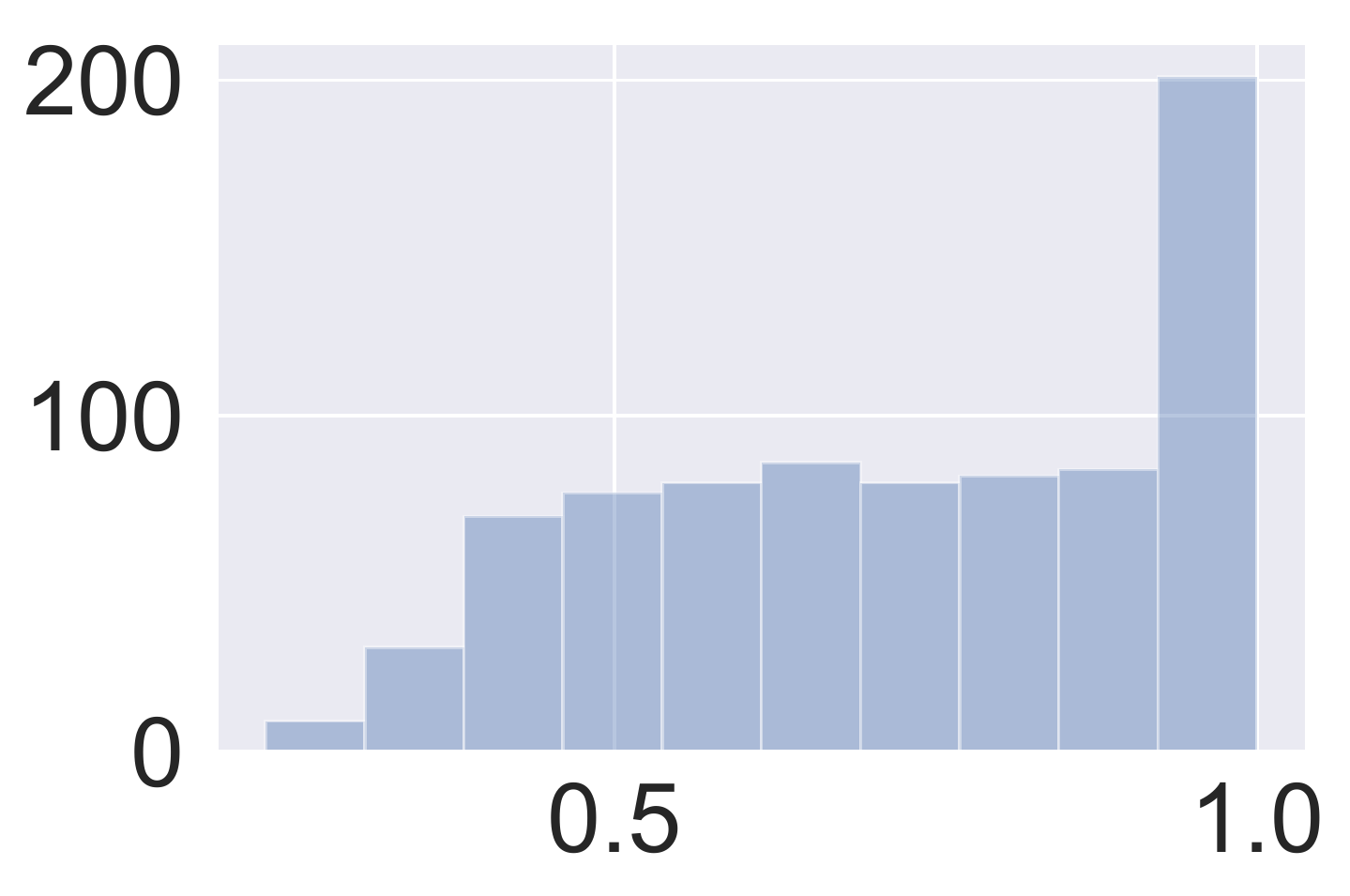}
		\caption{}
		\label{fig:dnn_monkey_ws_dist}
	\end{subfigure}
	\begin{subfigure}[b]{0.2\textwidth}
		\includegraphics[width=\columnwidth]{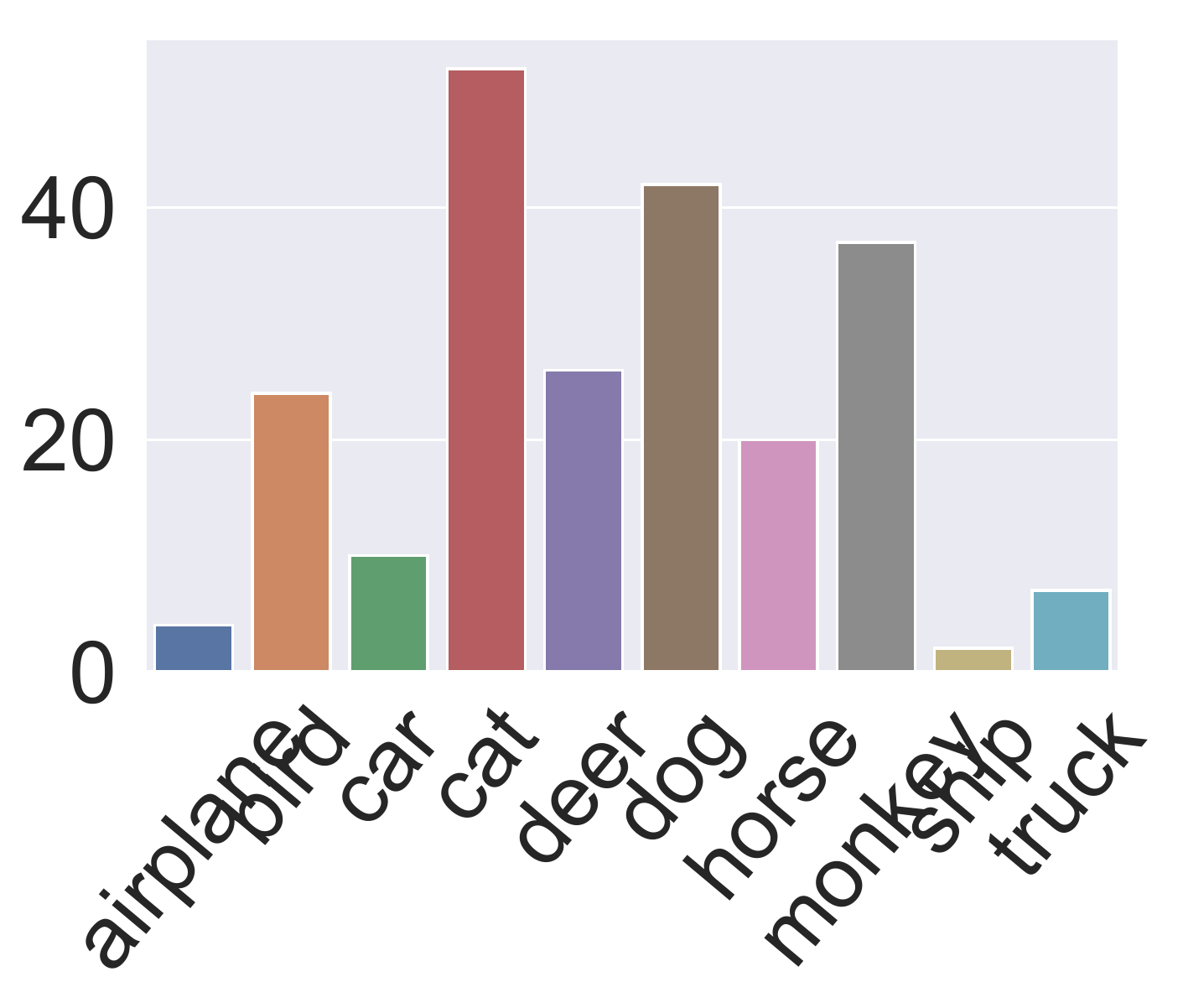}
		\caption{}
		\label{fig:dnn_monkey_high_ws_class_dist}
	\end{subfigure}
	\begin{subfigure}[b]{0.2\textwidth}
		\includegraphics[width=\columnwidth]{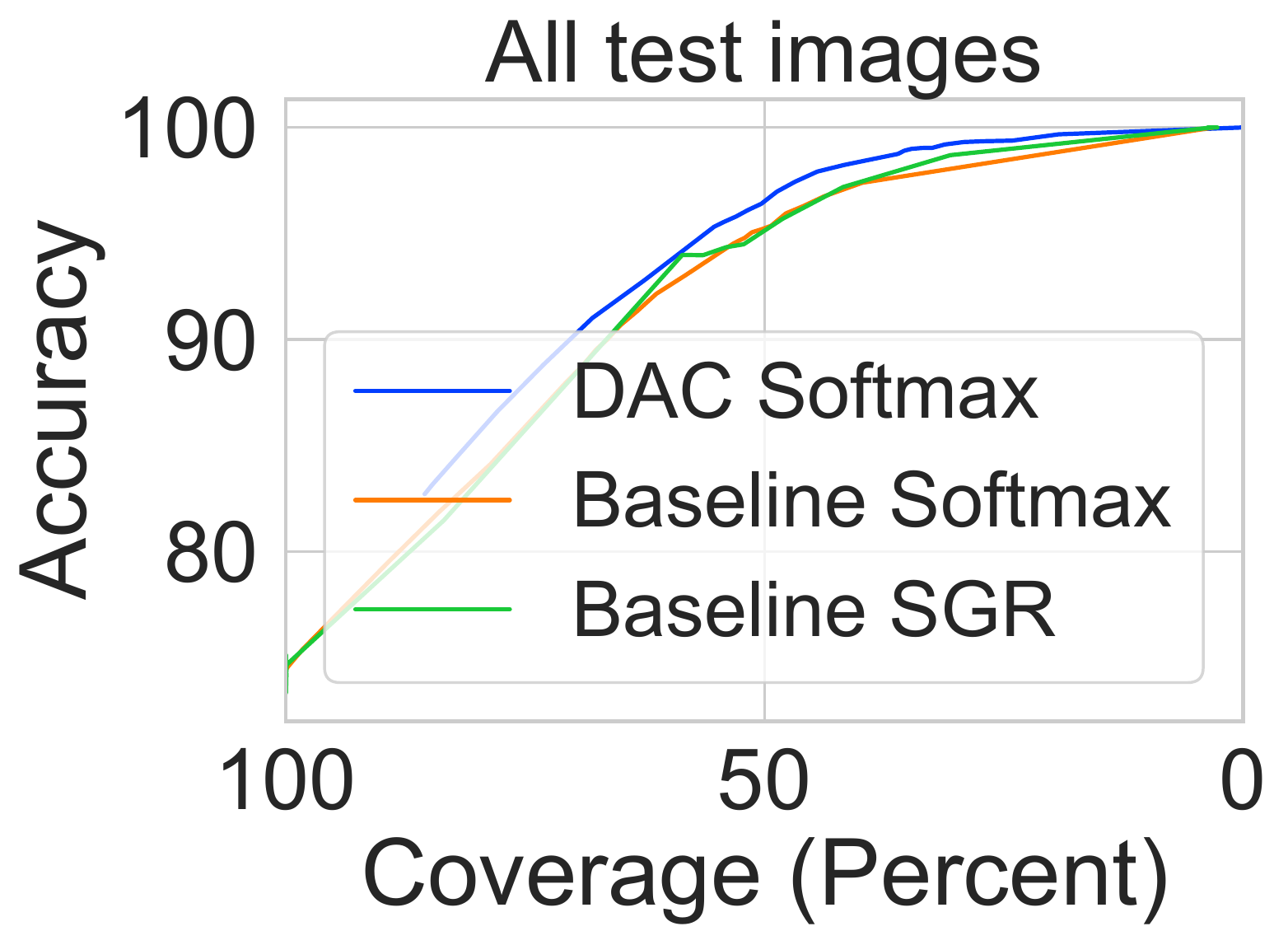}
		\caption{}
		\label{fig:rand_monkey_acc_vs_cov_all}
	\end{subfigure}
	\begin{subfigure}[b]{0.2\textwidth}
		\includegraphics[width=\columnwidth]{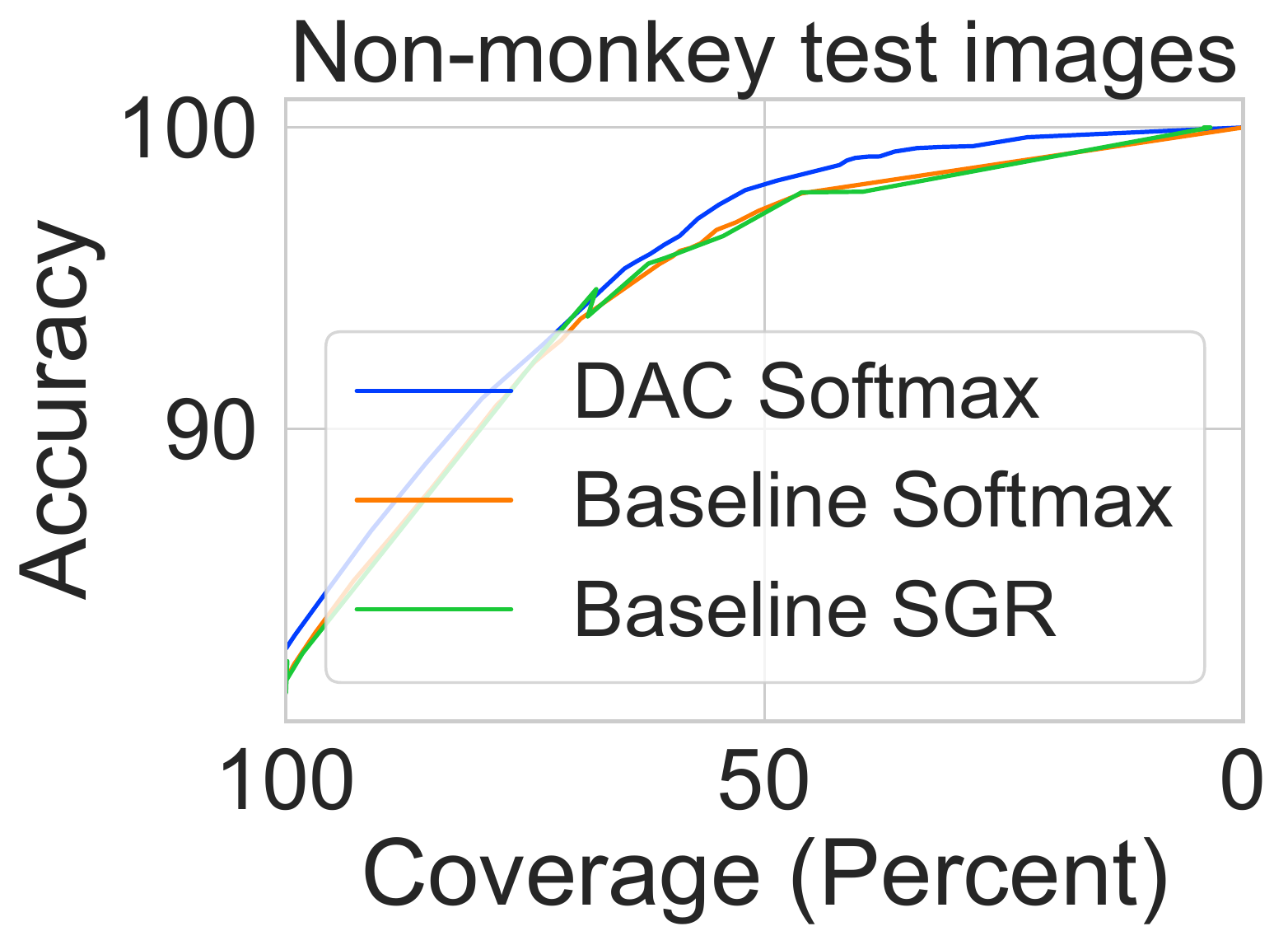}
		\caption{}
		\label{fig:rand_monkey_acc_vs_cov_na}
	\end{subfigure}

	\caption{(a) All monkey images in the training set had their labels randomized (b) The DAC abstains on about 80\% of the monkey images (abstention recall). (c) Among images that were abstained, most of the images were those of monkeys (abstention precision). (d) Distribution of baseline DNN predictions over monkey images in the test set indicating learning difficulty on this class (e) Distribution of winning softmax scores of the baseline DNN on the monkey images (f) Distribution of baseline DNN softmax scores > 0.9. Most of these confident predictions are non-monkeys (g) Comparison against various abstention methods (softmax and SGR) on all the test images (h) Same comparison, but only on those images on which the DAC did not abstain (i.e mostly non-monkey images). The DAC has a small but consistent advantage in both cases. All figures are computed on the test set.}
	\label{fig:random_monkey_results}
\end{figure*}

%
%

We first demonstrate the ability of the DAC to learn features associated with confusing labels in a toy experiment. In this experiment we simulate the  situation where an underlying, generally unknown feature occurring in a subset of the data often co-occurs with inconsistent mapping between features and ground truth. In a real-world setting, when encountering data containing such a feature, it is desired that the DAC will abstain on predicting on such samples and hand over the classification to an upstream (possibly human) expert. To simulate this, we randomize the labels (over the original $K$ classes) on 10\% of the images in the training set, but add a distinguishing extraneous feature to these images. In our experiments, this feature is a {\em smudge} (Figure~\ref{fig:smudged_sample}) that represents the aforementioned feature co-occurring  with label noise. We then train both a DAC as well as a regular DNN with the usual $K$-class cross-entropy loss. Performance is tested on a set where 10\% of the images are also smudged. Since it is hoped that the DAC learns representations for the structured noise occurring in the dataset, and assigns the abstention class for such training points, we also report the performance of a DNN that has been trained on a set where the abstained samples were eliminated ({\em post-DAC}) at the best-performing epoch of the DAC\footnote{we describe in Section~\ref{sec:unstructured_noise} how noisy data is eliminated} Performance is reported in terms of accuracy-vs-abstained (i.e., risk-vs-coverage) curves for the DAC, and the standard softmax threshold-based abstention for the baseline DNNs and post-DAC DNNs. As an additional baseline, we also compare the performance of the recently proposed selective guaranteed risk (SGR) method in ~\citep{geifman2017selective} for both the baseline and post-DAC DNNs that maximizes coverage subject to a user-specified risk bound (we use their default confidence parameter, $\delta$, of $0.001$ and report coverage  for a series of risk values.)


{\bf Results} When trained over a non-corrupted set, the baseline (i.e., non-abstaining) DNN had a test  accuracy over 82\%, but this drops to under 75\% (at 100\% coverage) when trained on the label-randomized smudged set (Figure~\ref{fig:smudge_dac_vs_dnn_softmax_threshold}).
For the DAC, Figure~\ref{fig:smudged_train_abstain} shows the abstention progress on the train set. When abstention is initiated  after 20 epochs, the DAC chooses to abstain on all but the easiest samples learned so far, but progressively abstains on less data till the abstention reaches steady behavior between epochs 120 and 150, abstaining on about 10\% of the data, representing the smudged images. Further annealing of learning rate (at epoch 160) causes the DAC to go into memorization mode.\footnote{We discuss abstention and memorization in Section~\ref{sec:abst_memorization}}. However, at the best performing validation epoch, the DAC abstains -- with both high precision and recall -- on precisely those set of images that have been smudged (Figures~\ref{fig:dac_smudged_abst_recall} and 
~\ref{fig:dac_smudge_pred_dist})! In other words, it appears the DAC has learned
%
%
%
 a clear association between the smudge  and unreliable training data, and opts to abstain whenever this feature is encountered in an image sample. Essentially, the smudge has become a separate class all unto itself, with the DAC assigning it the abstention class label. The risk-coverage curve for the DAC (Figure~\ref{fig:smudge_dac_vs_dnn_softmax_threshold}), calculated using softmax thresholds at the best validation epoch, closely tracks the baseline DNN's softmax thresholded curve; this is not surprising, since on those images that are not abstained on, the DAC and the DNN learn in similar fashion due to the way the loss function is constructed. We do however see a strong performance boost by eliminating the data abstained on by the DAC  and then re-training a DNN. This post-DAC  DNN has significantly higher accuracy than the baseline DNN (Figure~\ref{fig:smudge_dac_vs_dnn_softmax_threshold}), and also has consistently better risk-coverage curves. Not surprisingly this performance boost is also imparted to the SGR method since any improvement in the base accuracy of the classifier will be reflected in better risk-coverage curves. In this sense, the DAC is complementary to an  uncertainty quantification method like SGR or standard softmax thresholding -- first training with the DAC and then a DNN improves overall performance.
While this experiment clearly illustrates the DAC's ability to associate a particular feature with the abstention class, it might be argued the consistency of the smudge made this particular task easier than a typical real world setting. We provide a more challenging version of this experiment in the next section.

\subsection{Noisy Labels associated with a class}
\label{sec:rand_monkeys}

In this experiment, we simulate a scenario where a particular class, for some reason, is very prone to mislabeling, but it is assumed that given enough training data and clean labels, a deep network can learn the correct mapping. To simulate a rather extreme scenario, we randomize the labels over all the monkeys in the training set, which in fact include a variety of animals in the ape category (chimpanzees, macaques, baboons etc; Figure~\ref{fig:monkey_sample}) but all labeled as `monkey'. Unlike the previous experiment, where the smudge was a relatively simple and consistent feature, the set of features that the DAC now has to learn are over a complex real-world object with more intra-class variance. 
\label{subsec:rand_monkeys}


Detailed results are shown in Figure~\ref{fig:random_monkey_results}. The DAC  abstains on most of the monkey images in the test set (Figure ~\ref{fig:dac_monkey_abst_recall}), while abstaining on relatively many fewer images in the other classes (Figure ~\ref{fig:dac_monkey_abst_precision}), suggesting good abstention recall and precision respectively. In essence, the DAC, like a non-abstaining DNN would in a clean-data scenario, has learned meaningful representation of monkeys, but due to label randomization, the abstention loss function now encourages the DAC to associate monkey features with the abstention class. That is, the DAC, in the presence of label noise on this particular class, has learned a mapping from class features $X$ to class $K_{abstain}$, much like a regular DNN would have learned a mapping from $X$ to $K_{monkey}$ in the absence of label noise.  The representational power is unchanged from the DAC to the DNN; the difference is that the optimization induced by the loss function now redirects the mapping towards the abstention class.

Also shown is the performance of the baseline DNN in Figures~\ref{fig:dnn_monkey_dist} to~\ref{fig:dnn_monkey_high_ws_class_dist}. The prediction distribution over the monkey images spans the entire class range. That the DNN does get the classification correct about 20\% of the time is not surprising, given that about 10\% of the randomized monkey images did end up with the correct label, providing a consistent mapping from features to labels in these cases. However the accuracy  on monkey images is poor; the distribution of the winning softmax scores over the monkey images for the DNN is shown in 
Figure~\ref{fig:dnn_monkey_ws_dist}, revealing a high number of confident predictions ($p >= 0.9$) but  closer inspection of the class distributions across just these confident predictions (~\ref{fig:dnn_monkey_high_ws_class_dist}) reveals that most of these predictions are incorrect suggesting that a threshold-based approach, which generally works well~\citep{hendrycks2016baseline,geifman2017selective}, will produce confident but erroneous predictions in such cases. This is reflected in the small but consistent risk-vs-coverage advantage of the DAC in Figure~\ref{fig:rand_monkey_acc_vs_cov_all} and~\ref{fig:rand_monkey_acc_vs_cov_na}. As before we compare both a softmax-thresholded DAC and baseline DNN, as well as the SGR method tuned on the baseline DNN scores. Unlike the random smudging experiment, here we do not eliminate the abstained images and retrain -- doing so would completely eliminate one class. Instead we additionally compare the performance of the DAC on the images that it did not abstain (mostly non-monkeys), with the baselines (Figure~\ref{fig:rand_monkey_acc_vs_cov_na}) -- the DAC has a small but significant lead in this case as well.

In summary, the experiments in this section indicate that the DAC can reliably pick up and abstain on samples where the noise is correlated with an underlying feature. In the next section, we peek into the network for better insights into the features that cause the DAC to abstain.


\subsection{Visual Explanations of Abstention}
\begin{figure}[htbp]
	\centering
	\begin{subfigure}[b]{0.155\columnwidth}
		\includegraphics[width=\columnwidth]{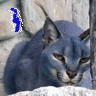}
		\caption{}
	\end{subfigure}
	\begin{subfigure}[b]{0.155\columnwidth}
		\includegraphics[width=\columnwidth]{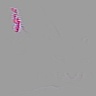}
		\caption{}
	\end{subfigure}
	\begin{subfigure}[b]{0.155\columnwidth}
		\includegraphics[width=\columnwidth]{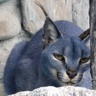}
		\caption{}
	\end{subfigure}
	\begin{subfigure}[b]{0.155\columnwidth}
		\includegraphics[width=\columnwidth]{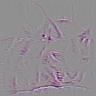}
		\caption{}
	\end{subfigure}
	\begin{subfigure}[b]{0.155\columnwidth}
		\includegraphics[width=\columnwidth]{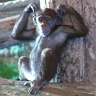}
		\caption{}
	\end{subfigure}
	\begin{subfigure}[b]{0.155\columnwidth}
		\includegraphics[width=\columnwidth]{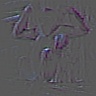}
		\caption{}
	\end{subfigure}
	\caption{Filter visualizations for the DAC. When presented with a smudged image (a), the smudge completely dominates the feature saliency map (b) that causes the DAC to abstain. However for the same image without a smudge (c), the class features become much more salient (d) resulting in a correct prediction. In the random monkeys experiment, for abstention on monkeys (e), the monkey features are picked up correctly (f), which leads to abstention}
	\label{fig:dac_vilter_viz}
\end{figure}

It is instructive to peer inside the network for explaining abstention behavior. Convolutional filter visualization techniques such as guided back-propagation~\citep{springenberg2014striving} combined with class-based activation maps~\citep{selvaraju2017grad} provide visually interpretable explanations of DNN predictions. In the case of the DAC, we visualize the final convolutional filters on the trained VGG-16 DAC model that successfully abstained on smudged and monkey images described in experiments in the previous section. Example visualizations using class-based activation maps on the predicted class are depicted in Figure~\ref{fig:dac_vilter_viz}. In the smudging experiments, when abstaining, the smudge completely dominates the rest of the features (Figures~\ref{fig:dac_vilter_viz}a,b). The same image, when presented without a smudge (Figure~\ref{fig:dac_vilter_viz}c), is correctly predicted, with the actual class features being much more salient (~\ref{fig:dac_vilter_viz}d) implying that the abstention decision is driven by the presence of the smudge.  For the randomized monkey experiment, it is precisely the features associated with the monkey class that result in abstention(Figures~\ref{fig:dac_vilter_viz}e, ~\ref{fig:dac_vilter_viz}f), visually confirming our hypothesis in Section~\ref{sec:rand_monkeys} that the DAC has effectively mapped monkey features to the abstention label.
Further experiments illustrating the abstention ability of the DAC in the presence of structured noise are described in Section B in the supplementary material. 


\section{Learning in the Presence of Unstructured Noise: The DAC as a Data Cleaner}
\label{sec:unstructured_noise}
\begin{table}[htbp]
	\centering
	\resizebox{\columnwidth}{!}{		
\begin{tabular}{l|l|llll}
	\hline
	\multicolumn{1}{c|}{\multirow{2}{*}{Dataset}}                                                & \multicolumn{1}{c|}{\multirow{2}{*}{Method}} & \multicolumn{4}{c}{Label Noise Fraction}                                                                                                                                                                                                                                                  \\
	\multicolumn{1}{c|}{}                                                                        & \multicolumn{1}{c|}{}                        & 0.2                                                                  & 0.4                                                                  & 0.6                                                                  & 0.8                                                                  \\ \hline
	\multirow{7}{*}{\begin{tabular}[c]{@{}l@{}}CIFAR-10\\ (ResNet-34)\end{tabular}}              & Baseline                                     & 88.94                                                                & 85.35                                                                & 79.74                                                                & 67.17                                                                \\
	& $\mathcal{L}_q$                              & 89.83                                                                & 87.13                                                                & 82.54                                                                & 64.07                                                                \\
	& Trunc $\mathcal{L}_q$                        & 89.7                                                                 & 87.62                                                                & 82.7                                                                 & 67.92                                                                \\
	& Forward $T$                                  & 88.63                                                                & 85.07                                                                & 79.12                                                                & 64.30                                                                \\
	& Forward $\hat{T}$                            & 87.99                                                                & 83.25                                                                & 74.96                                                                & 54.64                                                                \\
	& DAC                                          & \textbf{\begin{tabular}[c]{@{}l@{}}92.91\\ (0.24/0.01)\end{tabular}} & \textbf{\begin{tabular}[c]{@{}l@{}}90.71\\ (0.41/0.03)\end{tabular}} & \textbf{\begin{tabular}[c]{@{}l@{}}86.30\\ (0.56/0.07)\end{tabular}} & \textbf{\begin{tabular}[c]{@{}l@{}}74.84\\ (0.75/0.16)\end{tabular}} \\
	& Oracle                                       & 92.56                                                                & 90.95                                                                & 88.92                                                                & 86.43                                                                \\ \hline
	\multirow{4}{*}{\begin{tabular}[c]{@{}l@{}}CIFAR-10\\ (Wide Res-\\ Net 28x10)\end{tabular}}  & Baseline                                     & 91.53                                                                & 88.98                                                                & 82.69                                                                & 64.09                                                                \\
	& MentorNet                                    & 92.0                                                                 & 89.0                                                                 & -                                                                    & 49.0                                                                 \\
	& DAC                                          & \textbf{\begin{tabular}[c]{@{}l@{}}93.35\\ (0.25/0.01)\end{tabular}} & \textbf{\begin{tabular}[c]{@{}l@{}}90.93\\ (0.43/0.01)\end{tabular}} & \textbf{\begin{tabular}[c]{@{}l@{}}87.58\\ (0.59/0.04)\end{tabular}} & \textbf{\begin{tabular}[c]{@{}l@{}}70.8\\ (0.77/0.17)\end{tabular}}  \\
	& Oracle                                       & 95.17                                                                & 94.38                                                                & 92.74                                                                & 91.01                                                                \\ \hline
	\multirow{7}{*}{\begin{tabular}[c]{@{}l@{}}CIFAR-100\\ (ResNet-34)\end{tabular}}             & Baseline                                     & 69.15                                                                & 62.94                                                                & 55.39                                                                & 29.5                                                                 \\
	& $\mathcal{L}_q$                              & 66.81                                                                & 61.77                                                                & 53.16                                                                & 29.16                                                                \\
	& Trunc $\mathcal{L}_q$                        & 67.61                                                                & 62.64                                                                & 54.04                                                                & 29.60                                                                \\
	& Forward $T$                                  & 63.16                                                                & 54.65                                                                & 44.62                                                                & 24.83                                                                \\
	& Forward $\hat{T}$                            & 39.19                                                                & 31.05                                                                & 19.12                                                                & 8.99                                                                 \\
	& DAC                                          & \textbf{\begin{tabular}[c]{@{}l@{}}73.55\\ (0.18/0.05)\end{tabular}} & \textbf{\begin{tabular}[c]{@{}l@{}}66.92\\ (0.25/0.01)\end{tabular}} & \textbf{\begin{tabular}[c]{@{}l@{}}57.17\\ (0.77/0.03)\end{tabular}} & \textbf{\begin{tabular}[c]{@{}l@{}}32.16\\ (0.87/0.33)\end{tabular}} \\
	& Oracle                                       & 77.15                                                                & 73.85                                                                & 69.48                                                                & 58.5                                                                 \\ \hline
	\multirow{4}{*}{\begin{tabular}[c]{@{}l@{}}CIFAR-100\\ (Wide Res-\\ Net 28x10)\end{tabular}} & Baseline                                     & 71.24                                                                & 65.24                                                                & 57.56                                                                & 30.43                                                                \\
	& MentorNet                                    & 73.0                                                                 & 68.0                                                                 & -                                                                    & \textbf{35.0}                                                        \\
	& DAC                                          & \textbf{\begin{tabular}[c]{@{}l@{}}75.75\\ (0.2/0.05)\end{tabular}}  & \textbf{\begin{tabular}[c]{@{}l@{}}68.2\\ (0.57/0.01)\end{tabular}}  & \textbf{\begin{tabular}[c]{@{}l@{}}59.44\\ (0.76/0.06)\end{tabular}} & \begin{tabular}[c]{@{}l@{}}34.06\\ (0.87/0.33)\end{tabular}          \\
	& Oracle                                       & 78.76                                                                & 76.23                                                                & 72.11                                                                & 63.08                                                                \\ \hline
	\multirow{7}{*}{\begin{tabular}[c]{@{}l@{}}Fashion-\\ MNIST\\ (ResNet-18)\end{tabular}}      & Baseline                                     & 93.91                                                                & 93.09                                                                & 91.83                                                                & 88.61                                                                \\
	& $\mathcal{L}_q$                              & 93.35                                                                & 92.58                                                                & 91.3                                                                 & 88.01                                                                \\
	& $\mathcal{L}_q$                              & 93.21                                                                & 92.6                                                                 & 91.56                                                                & 88.33                                                                \\
	& Forward $T$                                  & 93.64                                                                & 92.69                                                                & 91.16                                                                & 87.59                                                                \\
	& Forward $\hat{T}$                            & 93.26                                                                & 92.24                                                                & 90.54                                                                & 85.57                                                                \\
	& DAC                                          & \textbf{\begin{tabular}[c]{@{}l@{}}94.76\\ (0.25/0.01)\end{tabular}} & \textbf{\begin{tabular}[c]{@{}l@{}}94.09\\ (0.48/0.01)\end{tabular}} & \textbf{\begin{tabular}[c]{@{}l@{}}92.97\\ (0.66/0.03)\end{tabular}} & \textbf{\begin{tabular}[c]{@{}l@{}}90.79\\ (0.88/0.04)\end{tabular}} \\
	& Oracle                                       & 95.22                                                                & 94.87                                                                & 94.64                                                                & 93.63                                                                \\ \hline
\end{tabular}

}
	\caption{Comparison of performance of DAC against related work for data corrupted with uniform label-noise. The DAC is used to first filter out noisy samples from the training set and a DNN is then trained on the cleaner set. Each set also shows the performance of the baseline DNN trained on the original data. Also shown is the performance of a hypothetical oracle data-cleaner that has perfect information about noisy labels. The parenthetical numbers next to the DAC indicate the fraction of training data removed by the DAC and the remaining noise level. 
		$\mathcal{L}_q$, truncated $\mathcal{L}_q$, and Forward results are from~\citep{zhang2018generalized}; MentorNet results are from ~\citep{jiang2018mentornet}} 
	\label{tab:label_rand_results}
\end{table}
So far we have seen the utility of the DAC in structured noise settings, where the DAC learns representations on which to abstain. Here we consider the problem of unstructured noise -- noisy labels that might occur arbitrarily on some fraction of the data. Classification performance degrades in the presence of noise~\citep{nettleton2010study}, with label noise shown to be more harmful than feature noise~\citep{zhu2004class}. While there have been a number of works related to DNN training in the presence of noise~\citep{sukhbaatar2014training,reed2014training,patrini2017making}, unlike these works we do not  model the label flipping probabilities between classes in detail. 
 We simply assume that a fraction of labels have been uniformly corrupted and  approach the problem from a data-cleaning perspective: using the abstention formulation and the extra class, can the DAC be used to identify noisy samples in the training set, with the goal of performing subsequent training, using a regular DNN, on the cleaner set? To identify the samples for elimination, we train the DAC, observing the performance of the non-abstaining part of the DAC on a validation set (which we assume to be clean). As mentioned before, this non-abstaining portion of the DAC 
 is simply the DAC with the abstention mass normalized out of the true classes.
 The result in Lemma 1 assures that learning continues on the true classes even in the presence of abstention. However at the point of best validation error, if there continues to be training error on the non-abstaining portion of the DAC, then this is likely indicative of label noise; it is these samples that are eliminated from the training set for subsequent training using regular cross-entropy loss.


To test the performance of the DAC, our comparisons include two recent models that achieve state-of-the-art results in training with noisy labels on image data: MentorNet~\citep{jiang2018mentornet} that uses a data-driven curriculum-learning approach involving two neural nets -- a learning network (StudentNet) and a supervisory network (MentorNet); and~\citep{zhang2018generalized}, that uses a noise-robust loss function formulation involving a generalization of the traditional categorical cross-entropy loss function. We also compare the performance against the Forward method described in~\citep{patrini2017making} which uses a loss correction approach; for the latter we use the numbers reported in~\citep{zhang2018generalized} using the same setup.

{\bf Experimental Setup} We use the CIFAR-10, CIFAR-100~\citep{krizhevsky2009learning} and the Fashion-MNIST ~\citep{xiao2017fashion} datasets with an increasing fraction of arbitrarily randomized labels, using the same networks  as the ones we compare to. In the DAC approach, both the DAC and the downstream DNN (that trains on the cleaner set) use the same network architectures. 
The downstream DNN is trained using the same hyperparameters, optimization algorithm and weight decay as the models we compare to. As a best-case model in the data-cleaning scenario, we also report the performance of a hypothetical oracle that has perfect information of the corrupted labels, and eliminates only those samples. To ensure approximately the same number of optimization steps as the comparisons when data has been eliminated, we appropriately lengthen the number of epochs and learning rate schedule for the downstream DNN (and do the same for the oracle.)

Results are shown in Table~\ref{tab:label_rand_results}. By identifying and eliminating noisy samples using the DAC and then training using the cleaner set,  noticeable -- and often significant --  performance improvement is achieved over the comparison methods in most cases. Interestingly, in the case of higher label randomization, for the more challenging data set like CIFAR-10 and CIFAR-100, we see the noisy baseline outperforming some of the comparison methods. The DAC is however, consistently better than the baseline. On CIFAR-100, for 80\% randomization, the other methods often have very similar performance to the DAC. This is possibly due to substantial the amount of data that has been eliminated by the DAC leaving very few samples per class. The fact that the performance is comparable even in this case, and the high hypothetical performance of the oracle illustrate the effectiveness of a data cleaning approach for deep learning  even when a significant fraction of the  data has been eliminated. Additional results in the case of non-uniform, class-dependent label noise are reported in Section C of the Appendix.


While data cleaning (or pruning) approaches have been considered before in the context of shallow classifiers~\citep{angelova2005pruning,brodley1999identifying,zhu2003eliminating}, to the best of our knowledge, this is the first work to show how abstention training can be used to identify and eliminate noisy labels for improving classification performance. Besides the improvements over the work we compare to, this approach also has additional advantages: we do not need to estimate the label confusion matrix as in ~\cite{sukhbaatar2014training,reed2014training,patrini2017making} or make assumptions regarding the amount of label noise or the existence of a trusted or clean data set as done in~\citep{hendrycks2018using} and ~\citep{li2017learning}.

The DAC approach is also significantly simpler than methods based on the mentor-student networks in~\citep{jiang2018mentornet,han2018co}, or the graphical model approach in~\citep{vahdat2017toward}.  The results in this section not only demonstrate the performance benefit of a data-cleaning approach for robust deep learning in the presence of significant label noise, but also the utility of the abstaining classifier as an effective way to clean such noise.

\section{Abstention and Memorization}
\label{sec:abst_memorization}


%
%

In the structured-noise experiments in Section~\ref{sec:struct_noise}, we saw that the DAC abstains, often with near perfection, on label-randomized samples by learning common features that are present in these samples. However, there has been a considerable body of recent work that shows that DNNs are also perfectly capable of memorizing random labels~\citep{zhang2016understanding,arpit2017closer}. In this regard, abstention appears to counter the tendency to memorize data; however it does not generally prevent memorization. 

\begin{lem}
	For the loss function $\mathcal{L}$ given in Equation~\ref{eqn:loss_fn}, for a fixed $\alpha$, and trained over $t$ epochs, as $t \rightarrow \infty$, the abstention rate $\gamma \rightarrow 0$ or $\gamma \rightarrow 1$.
\end{lem}

\begin{figure}
	\begin{subfigure}{0.45\columnwidth}
		\includegraphics[scale=0.25]{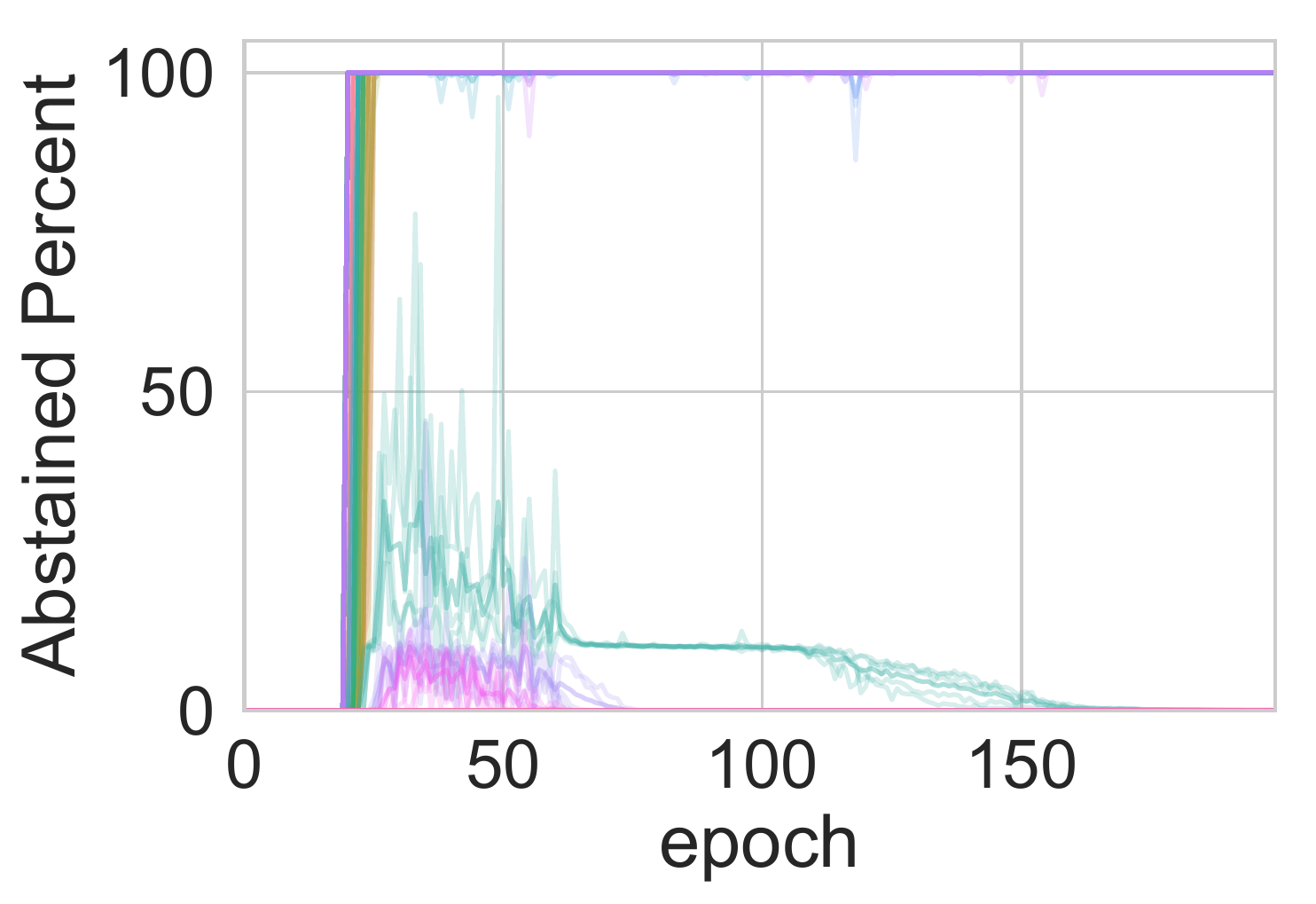}
		\label{fig:abst_behavior_alpha}
	\end{subfigure}
	\begin{subfigure}{0.45\columnwidth}
		\includegraphics[scale=0.25]{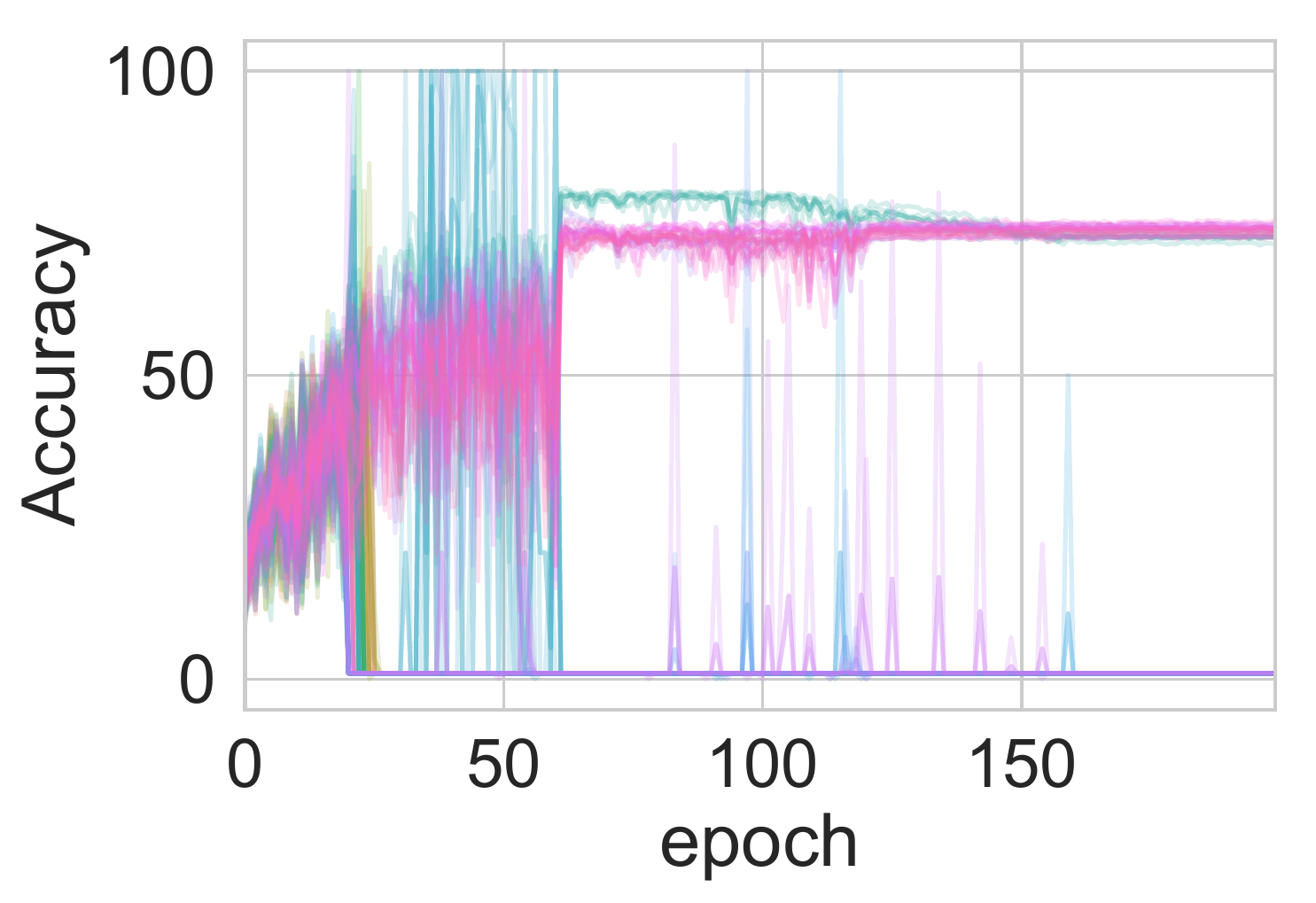}
		\label{fig:val_behavior_alpha}
	\end{subfigure}
	\vspace{-0.18in}
	\caption{DAC behavior for the random smudging experiments for different fixed $\alpha$'s (indicated by color) showing the tendency for all-or-nothing abstention (left) if $\alpha$ is kept constant. In the zero-abstention case, accuracy is the same as a baseline non-abstaining DNN (right)} 
	
	\label{fig:alpha_behavior}
\end{figure}
\vspace{-0.1in}
\begin{sproof}
 Intuitively, if $\alpha$ is close to 0, $p_{k+1}$ quickly saturates to unity, causing the DAC to abstain on all samples, driving both loss and the gradients to $0$ and preventing any further learning. Barring this situation, and given that the gradient $\frac{\partial \mathcal{L} }{\partial a_{j}} \leq 0$, where $j$ is the true class (see Lemma 1), the condition for abstention in Section 2 eventually fails to be satisfied. After this point, probability mass is removed from the abstention class ${k+1}$ for all subsequent training, eventually driving abstention to zero. 
 \end{sproof}
\vspace{-0.1in}

Experiments where $\alpha$ was fixed confirm this; Figure~\ref{fig:alpha_behavior} shows abstention behavior and the corresponding generalization performance on the validation set for different values of fixed alpha in the random-smudging experiments (Section~\ref{sec:rand_smudge}). The desired behavior of abstaining on the smudged samples (whose labels were randomized in the training set) does not persist indefinitely. At epochs 60 and 120, there are steep reductions in the abstention rate, coinciding with learning rate decay. At this point, apparently, the DAC moves into a memorization phase, finding more complex decision boundaries to fit random labels, as the lower learning rate enables it to descend into a possibly narrow minima. Generalization performance also suffers once this phase begins. This behavior is consistent with the discussion in~\citep{arpit2017closer} -- the DAC does indeed first learn patterns before descending into memorization. 
 Auto-tuning of $\alpha$ described in Section~\ref{sec:alpha_auto} prevents the abstention rate from saturating to 1; however as we saw in Section~\ref{sec:rand_smudge}, it does not prevent abstention from converging to 0. A sufficiently small learning rate and long training schedule eventually results in memorization. As discussed earlier, tracking the loss of the non-abstaining portion of the DAC on a validation set can used to  determine when a desirable abstention level has been reached.
\section{Conclusions}
\label{sec:conclusion}

There is little work discussing abstention in the context of deep learning and even less discussing abstention approaches for combating label noise. Here, we  demonstrated the effectiveness of such an  approach when training deep neural networks. We showed the utility of the DAC under multiple types of label noise: as a representation learner in the presence of structured noise and as an effective data cleaner in the presence of arbitrary noise. 
  Results indicate that data-cleaning with the DAC significantly improves classification performance  for downstream training. Furthermore, the loss function formulation is simple to implement and  can work with any existing DNN architecture; this makes the DAC a useful addition to real-world deep learning pipelines. 

{\bf Acknowledgments:} The authors were supported in part by the Joint Design of Advanced Computing Solutions for Cancer (JDACS4C) program established by the U.S. Department of Energy (DOE) and the National Cancer Institute (NCI) of the National Institutes of Health. This work was performed under the auspices of the U.S. Department of Energy by  Los Alamos National Laboratory under Contract DE-AC5206NA25396. This work was supported in part by the CONIX Research Center, one of six centers in JUMP, a Semiconductor Research Corporation (SRC) program sponsored by DARPA.
%

\bibliography{deepuq}
\bibliographystyle{icml2019}

\end{document}


\begin{appendices}
		\setcounter{equation}{0}
		\section{Proof of Lemma 1}
		\label{app:learning_proof}
		
		Here we show that even in the presence of abstention, learning continues on the true classes. 				Consider again the loss function defined 
		for a sample $x$.
		\small
		\begin{equation*}
		\begin{aligned}
		\mathcal{L}(x) =
		( 1 - p_{k+1})\left(-\sum_{i=1}^k t_i \log \frac{p_i}{1-p_{k+1}}\right)
		+ \alpha \log \frac{1} { 1 - p_{k+1} }.
		\end{aligned}
		\end{equation*}
		\normalsize
		
		Let $j$, ($1 \leq j \leq k$) be the true class for $x$. During gradient descent, learning on the true class takes place if $\frac{\partial \mathcal{L} }{\partial a_{j}} < 0$, where $a_j$ is the pre-activation into the softmax unit of class $j$.  
		
		A straight-forward gradient calculation shows that 
		\small
		\begin{equation*}
		\begin{aligned}
		\frac{\partial \mathcal{L} }{\partial a_{j}} &= -(1- p_j - p_{k+1}) + p_{k+1}p_j\log\left({\frac{1-p_{k+1}}{p_j}}\right) \\ &- \alpha \frac{p_{k+1}p_j}{1-p_{k+1}}
		\end{aligned}
		\end{equation*}
		\normalsize
		Since $\alpha \geq 0$ as per as our assumption, the last quantity in the above expression, $- \alpha \frac{p_{k+1}p_j}{1-p_{k+1}} \leq 0$
		
		Also note that $(1-p_j-p_{k+1})$ in the above expression is just the total probability mass in the remaining real (i.e., non-abstention) classes; denote this by $q$.
		
		Then we have
		\begin{equation*}
		\begin{aligned}
		-(1- p_j - p_{k+1}) + p_{k+1}p_j\log\left({\frac{1-p_{k+1}}{p_j}}\right) \\= -q + p_{k+1}p_j\log\left({\frac{1-p_{k+1}}{p_j}}\right) \\
		= -q + p_{k+1}p_j\log\left({\frac{q+p_j}{p_j}}\right) \\
		= -q + p_{k+1}p_j\log\left({1+\frac{q}{p_j}}\right) \\
		\leq -q + p_{k+1}p_j\frac{q}{p_j} \label{eqn:log_ineq_follows} \\ 
		= -q + p_{k+1}q \\
		\leq 0
		\end{aligned}
		\end{equation*}
		
		where, in ~\ref{eqn:log_ineq_follows}, we have made use of the fact that $\log\left(1 + x\right) \leq x$ for all  $x > -1$. Thus $\frac{\partial \mathcal{L} }{\partial a_{j}} \leq 0$ as desired.

		\section{Noisy Labels associated with a data transformation}
		\label{app:blur_expts}
\begin{figure*}
			\centering
			\begin{subfigure}[b]{0.35\textwidth}
				\includegraphics[width=\columnwidth]{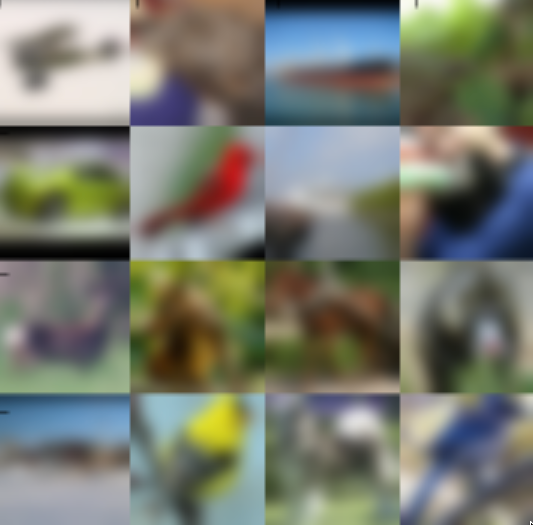}
				\caption{}
				\label{fig:blurred_sample}
			\end{subfigure}
			\hspace{0.1in}
			\begin{subfigure}[b]{0.45\textwidth}
				\includegraphics[width=\columnwidth]{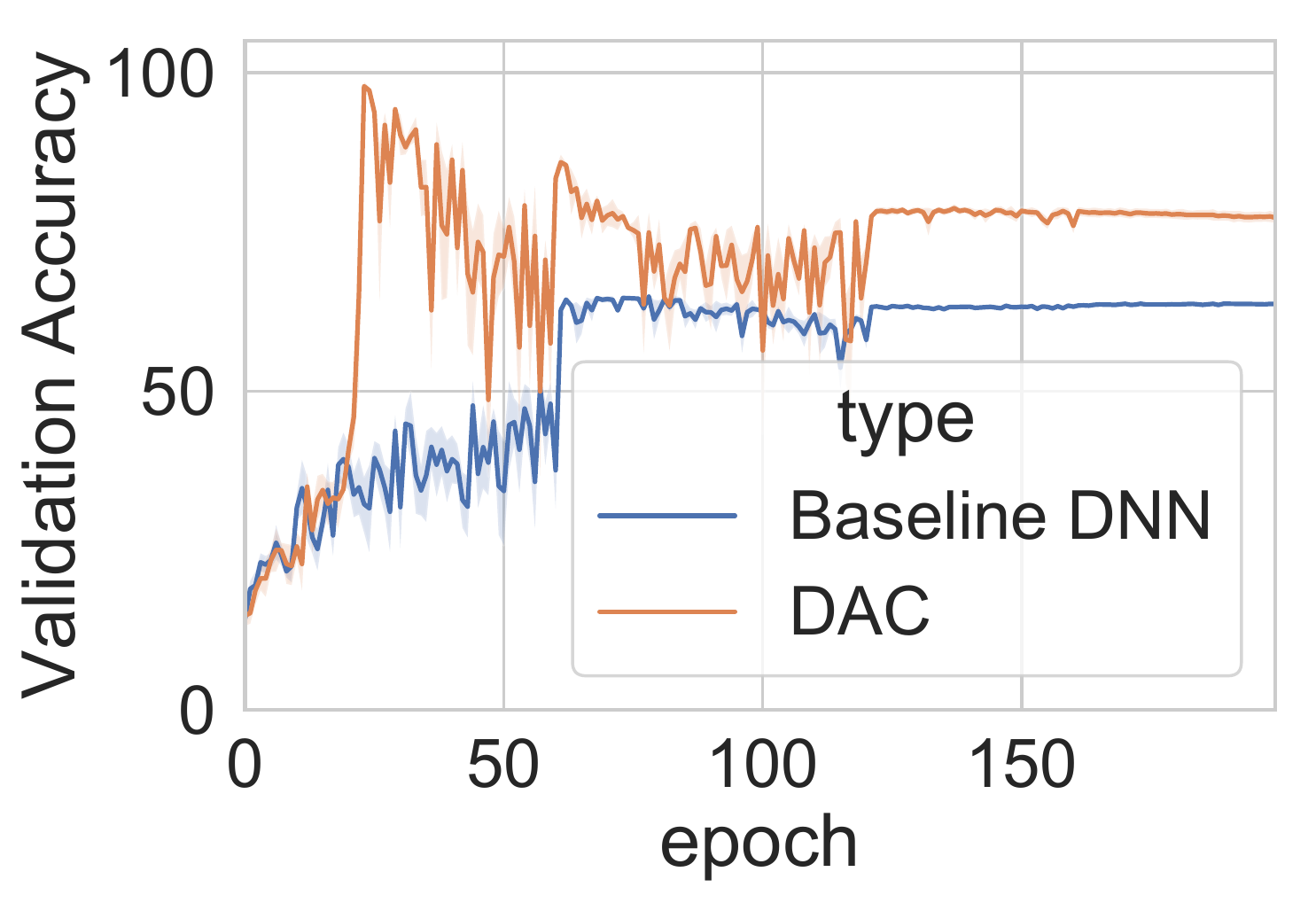}
				\caption{}
				\label{fig:blurred_dnn_vs_dac}
			\end{subfigure}
			\hspace{0.1in}
			\begin{subfigure}[b]{0.45\textwidth}
				\includegraphics[width=\columnwidth]{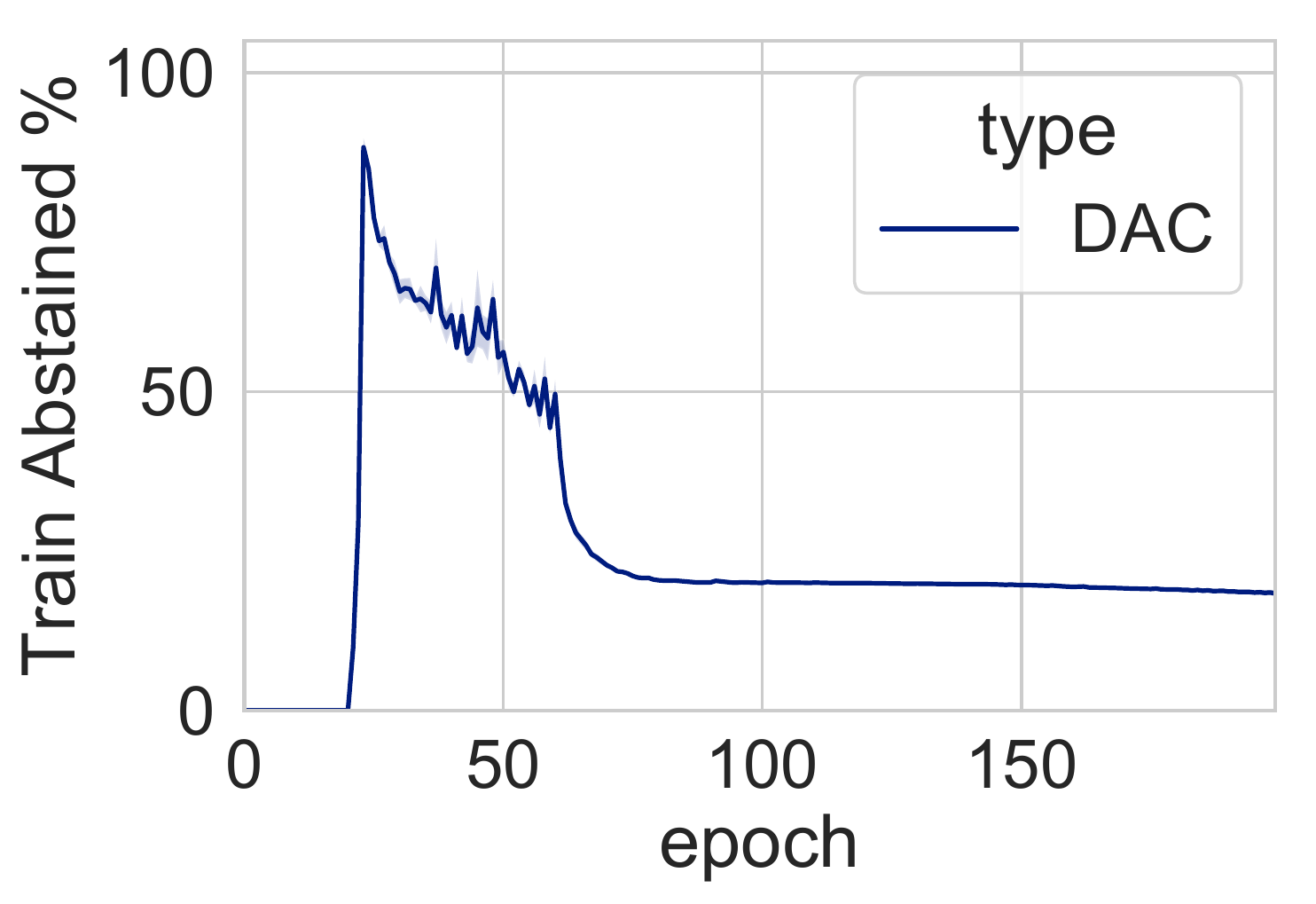}
				\caption{}
				\label{fig:blurred_abst_epoch}
			\end{subfigure}
			\hspace{0.1in}
			\begin{subfigure}[b]{0.45\textwidth}
				\includegraphics[width=\columnwidth]{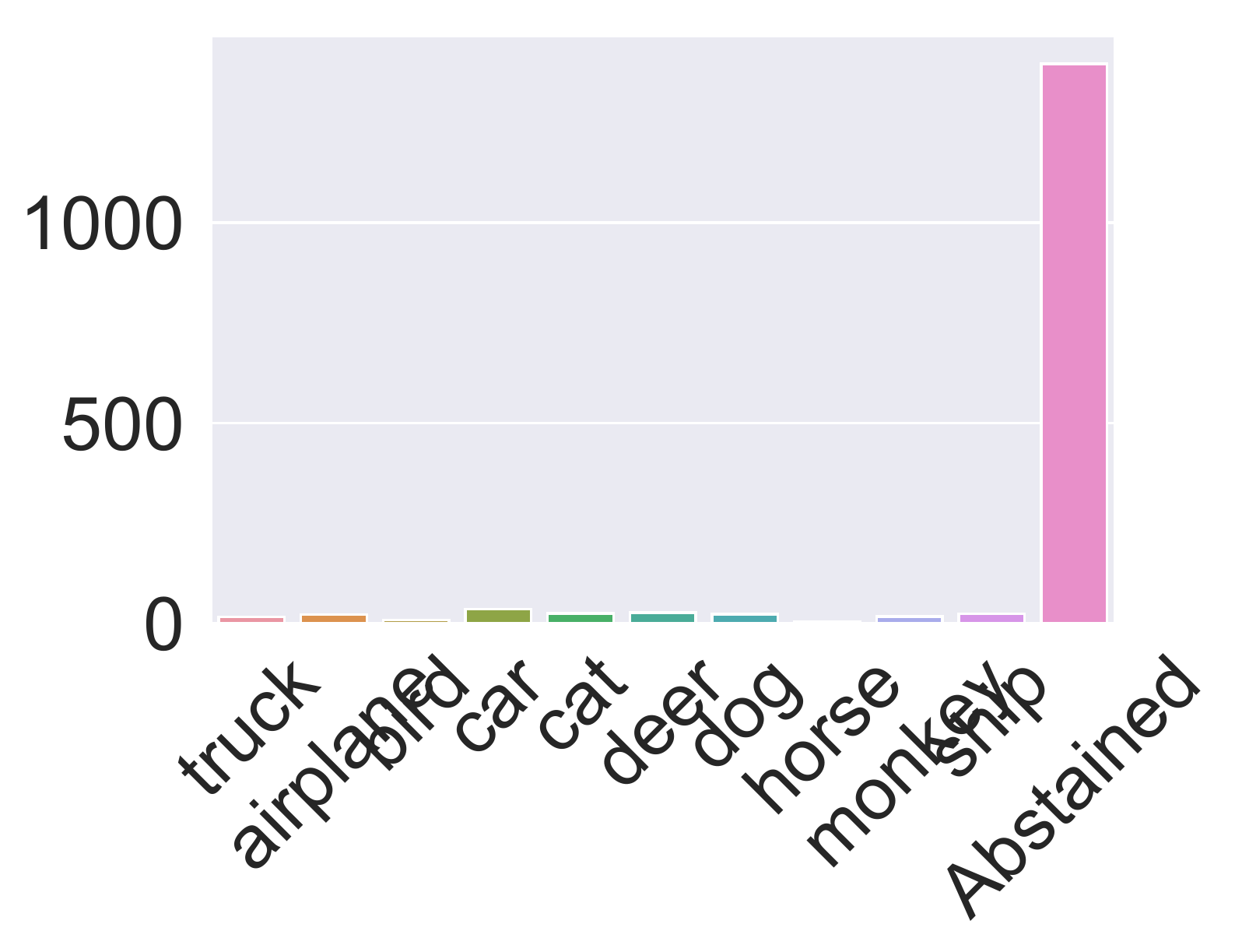}
				\caption{}
				\label{fig:blurred_pred_dist}
			\end{subfigure}
			\caption{Results on blurred-image experiment with noisy labels (a)20\% of the images are blurred in the train set, and their labels randomized (b) Validation accuracy for baseline vs DAC (non-abstained) (c)Abstention behavior for the DAC during training (d) Distribution of predictions on the blurred validation images for the DAC. We also observed (not shown) that for the baseline DNN, the accuracy on the blurred images in the validation set is no better than random.}
			\label{fig:blur_results}
		\end{figure*}

		We present further results on the abstaining ability of the DAC in the presence of structured noise. Here we simulate a scenario where a subset of the training data, due to feature degradation, ends up with unreliable labels. We apply a Gaussian blurring transformation to 20\% of the train and test images across all the classes (Figure~\ref{fig:blurred_sample}), and randomize the labels on the blurred training set. This is similar to the smudging experiment, but lacks the presence of a consistent, conspicuous feature that the DAC can associate with abstention. On the other hand, the lack of high frequency components, or conversely the abundance of low frequency components, might itself be thought of as a feature that is consistent across the samples that have had their label randomized.
		{\bf Results} The DAC abstains remarkably well on the blurred images in the test set (Figure~\ref{fig:blurred_pred_dist}), while maintaining classification accuracy over the remaining samples in the validation set ($\approx$ 79\%). The baseline DNN accuracy drops to 63\% (Figure~\ref{fig:blurred_dnn_vs_dac}), while the baseline accuracy over the smudged images alone is no better than random ($\approx 9.8\%$) . The abstention behavior of the DAC on the blurred images in the test set can be explained by how abstention evolves during training (Figure~\ref{fig:blurred_abst_epoch}). Once abstention is introduced at epoch 20, the DAC initially opts to abstain on a high percentage of the training data, while continuing to learn (since the gradients w.r.t the true-class pre-activations are always negative.). In the later epochs, sufficient learning has taken place on the non-randomized samples but the DAC continues to abstain on about 20\% of the training data, which corresponds to the blurred images indicating that a strong association has been made between blurring and abstention.

		\section{Results on Non-Uniform Label Noise}
		\begin{table}[!htb]
			\begin{tabular}{c|l|llll}
				\hline
				\multirow{2}{*}{Dataset}                                                                & \multicolumn{1}{c|}{\multirow{2}{*}{Method}} & \multicolumn{4}{c}{\begin{tabular}[c]{@{}c@{}}Class Dependent\\  Label Noise Fraction\end{tabular}} \\
				& \multicolumn{1}{c|}{}                        & $\eta=$0.1                     & 0.2                     & 0.3                    & 0.4                    \\ \hline
				\multirow{5}{*}{\begin{tabular}[c]{@{}c@{}}CIFAR-10\\ (ResNet-34)\end{tabular}}         & $\mathcal{L}_q$                              & 90.91                   & 89.33                   & 85.45                  & 76.74                  \\
				& Trunc $\mathcal{L}_q$                        & 90.43                   & 89.45                   & 87.10                  & 82.28                  \\
				& Forward $T$                                  & 91.32                   & 90.35                   & 89.25                  & 88.12                  \\
				& Forward $\hat{T}$                            & 90.52                   & 89.09                   & 86.79                  & 83.55                  \\
				& DAC                                          & \textbf{94.23}          & \textbf{93.20}          & \textbf{92.07}         & \textbf{89.88}         \\ \hline
				\multirow{5}{*}{\begin{tabular}[c]{@{}c@{}}CIFAR-100\\ (ResNet-34)\end{tabular}}        & $\mathcal{L}_q$                              & 68.36                   & 66.59                   & 61.45                  & 47.22                  \\
				& Trunc $\mathcal{L}_q$                        & 68.86                   & 66.59                   & 61.87                  & 47.66                  \\
				& Forward $T$                                  & 71.05                   & 71.08                   & 70.76                  & \textbf{70.82}                  \\
				& Forward $\hat{T}$                            & 45.96                   & 42.46                   & 38.13                  & 34.44                  \\
				& DAC                                          & \textbf{75.59}          & \textbf{73.22}          & \textbf{71.38}         & 65.34         \\ \hline
				\multirow{5}{*}{\begin{tabular}[c]{@{}c@{}}Fashion-MNIST\\ \\ (ResNet-18)\end{tabular}} & $\mathcal{L}_q$                              & 93.51                   & 93.24                   & 92.21                  & 89.53                  \\
				& Trunc $\mathcal{L}_q$                        & 93.53                   & 93.36                   & 92.76                  & 91.62                  \\
				& Forward $T$                                  & 94.33                   & 94.03                   & 93.91                  & 93.65                  \\
				& Forward $\hat{T}$                            & 94.09                   & 93.66                   & 93.52                  & 88.53                  \\
				& DAC                                          & \textbf{95.48}          & \textbf{95.08}          & \textbf{94.96}         & \textbf{94.31}         \\ \hline
			\end{tabular}
			\caption{Comparison of DAC vs related methods for class-dependent label noise. Performance numbers reproduced from ~\citep{zhang2018generalized}. For the DAC, an abstaining classifier is first used to identify and eliminate label noise, and an identical DNN is then used for downstream training.}
			\label{tab:non_uniform_noise_results}
		\end{table}
		Here we report results on CIFAR-10, CIFAR-100 and Fashion-MNIST for class-dependent label noise. The experimental setup is exactly as described in ~\citep{zhang2018generalized}, and we compare with the results reported in that paper which also includes the Forward correction method of~\citep{patrini2017making}. In CIFAR-10, the class dependent noise results in the following flip scenario with probability $\eta$:  TRUCK $\rightarrow$ AUTOMOBILE, BIRD $\rightarrow$ AIRPLANE, DEER $\rightarrow$ HORSE, and CAT $\leftrightarrow$ DOG with probability. For CIFAR-100,  classes are organized into groups as described in~\citep{krizhevsky2009learning} and the  class-dependent noise is simulated by flipping each class into the next circularly with probability $\eta$. For Fashion-MNIST, classes are flipped as follows: BOOT $\rightarrow$ SNEAKER , SNEAKER $\rightarrow$ SANDALS, PULLOVER $\rightarrow$ SHIRT, COAT$\rightarrow$ DRESS with probability $\eta$. The aforementioned flipping scenarios are identical to the setup in ~\citep{zhang2018generalized}. Results are shown in Table~\ref{tab:non_uniform_noise_results}.



	\end{appendices}
	
	\bibliography{deepuq}
	\bibliographystyle{icml2019}